\documentclass[conference]{IEEEtran}
\IEEEoverridecommandlockouts
\usepackage[switch]{lineno} 

\usepackage[utf8]{inputenc} 
\usepackage[T1]{fontenc}    
\usepackage{booktabs}       
\usepackage{amsfonts}       
\usepackage{nicefrac}       
\usepackage{xcolor}         
\usepackage{color}
\usepackage{ulem}

\usepackage{url}

\usepackage{booktabs} 

\usepackage{multirow}
\usepackage{subcaption}

\usepackage{enumitem}

\usepackage{wrapfig}

\usepackage{collcell}
\usepackage{xstring}
\usepackage{datatool}
\usepackage{booktabs}
\usepackage{environ}

\usepackage{color}
\definecolor{darkblue}{rgb}{0.0, 0.2, 0.6}

\usepackage{balance}
\usepackage{float}
\usepackage{listings}
\usepackage{color}
\definecolor{dkgreen}{rgb}{0,0.6,0}
\definecolor{gray}{rgb}{0.5,0.5,0.5}
\definecolor{mauve}{rgb}{0.58,0,0.82}
\definecolor{bg}{rgb}{0.9,0.9,0.9}

\newcounter{CurrentRow}
\newcounter{CurrentColumn}
\newtoggle{DoneWithFirstRow}

\newcommand*{\FirstColumn}[1]{%
    \IfEq{\arabic{CurrentColumn}}{0}{%
        \global\togglefalse{DoneWithFirstRow}%
        \setcounter{CurrentRow}{1}
    }{%
        \global\toggletrue{DoneWithFirstRow}%
        \stepcounter{CurrentRow}%
    }%
    \setcounter{CurrentColumn}{0}%
    \NewData{#1}%
}
\newcommand*{\NewData}[1]{%
    \dtlexpandnewvalue%
    \stepcounter{CurrentColumn}%
    \iftoggle{DoneWithFirstRow}{%
        \dtlgetrow{TransposedTabularDB}{\arabic{CurrentColumn}}%
        \dtlappendentrytocurrentrow{\Alph{CurrentRow}}{#1}%
        \dtlrecombine%
    }{%
        \DTLnewrow{TransposedTabularDB}%
        \DTLnewdbentry{TransposedTabularDB}{\Alph{CurrentRow}}{#1}%
    }%
}%

\newcolumntype{F}{>{\collectcell\FirstColumn}c<{\endcollectcell}}
\newcolumntype{C}{>{\collectcell\NewData}{c}<{\endcollectcell}}

\newcommand{\twomm}{\textsc{2mm}\xspace}

\newtoggle{EncounteredDataRow}

\newsavebox{\TempBox}
\DTLnewdb{TransposedTabularDB}

\NewEnviron{Ttabular}[1]{%
    \setcounter{CurrentColumn}{0}%
    \global\togglefalse{EncounteredDataRow}%
    \savebox{\TempBox}{%
        \begin{tabular}{FCCCCCC}
            \BODY%
        \end{tabular}%
    }%
    \begin{tabular}{#1}\toprule%
    \DTLforeach*{TransposedTabularDB}{\Aa=A, \Ba=B, \Ca=C}{%
      \DTLiffirstrow{}{\\\midrule}%
        \Aa & \Ba & \Ca %
    }\\\bottomrule%
    \end{tabular}%
}%

\usepackage[colorinlistoftodos]{todonotes}

\definecolor{applegreen}{rgb}{0.55, 0.71, 0.0}

\usepackage{twoopt}
\usepackage{tikz}
\usetikzlibrary{fit}

\newcommandtwoopt\Textbox[5][2.5cm][2cm]{%
\begin{tikzpicture}[remember picture,overlay]
  \coordinate (aux) at ([xshift=#1]#4);
  \node[inner ysep=3pt,yshift=0.6ex,draw=green,thick,
    fit=(#3) (aux),baseline] 
    (box) {};
  \node[text width=#2,anchor=north east,
    font=\sffamily\footnotesize,align=right] 
    at (box.north east) {#5};
\end{tikzpicture}%
}
\DeclareCaptionLabelFormat{alglabel}{Alg.~#2}

\usepackage{xspace}

\newcommand{\corr}{\textsc{correlation}\xspace}

\newcommand{\cov}{\textsc{covariance}\xspace}

\newcommand{\gemmNcubed}{\textsc{Gemm-n}\xspace}

\newcommand{\trmmOpt}{\textsc{trmm-opt}\xspace}

\newcommand{\programl}{\textsc{ProGraML}\xspace}
\newcommand{\codetfive}{\textsc{CodeT5}\xspace}
\newcommand{\greaseLM}{\textsc{GreaseLM}\xspace}
\newcommand{\tfconv}{\textsc{TransformerConv}\xspace}

\newcommand{\codeTwoVec}{\textsc{Code2vec}\xspace}

\newcommand{\cdfg}{\textsc{HARP}\xspace}
\newcommand{\simpleConcat}{\textsc{ProgSG-ca}\xspace}
\newcommand{\sumInt}{\textsc{ProgSG-si}\xspace}

\newcommand{\model}{\textsc{ProgSG}\xspace}

\newcommand{\nop}[1]{}

\usepackage{natbib}
\usepackage{amsmath}
\usepackage{amsthm}
\usepackage{bm}
\usepackage{algorithm}
\usepackage{algorithmic}
\usepackage{bbm}

\usepackage{mathtools}

\usepackage[capitalize,noabbrev]{cleveref}

\theoremstyle{plain}

\theoremstyle{definition}

\theoremstyle{remark}

\def\BibTeX{{\rm B\kern-.05em{\sc i\kern-.025em b}\kern-.08em
    T\kern-.1667em\lower.7ex\hbox{E}\kern-.125emX}}
\begin{document}


\title{Cross-Modality Program Representation Learning for Electronic Design Automation with High-Level Synthesis
}

\author{\IEEEauthorblockN{Zongyue Qin\textsuperscript{*}, Yunsheng Bai\textsuperscript{*}, Atefeh Sohrabizadeh, Zijian Ding, Ziniu Hu, Yizhou Sun, Jason Cong}
\IEEEauthorblockA{\textit{Computer Science} \\
\textit{UCLA}\\
Los Angeles, US\\
\{qinzongyue,yba,atefehsz,bradyd,bull,yzsun,cong\}@cs.ucla.edu}\thanks{\textsuperscript{*}The two authors contributed equally to this work.}
}

\maketitle

\begin{abstract}


In recent years, domain-specific accelerators (DSAs) have gained popularity for applications such as deep learning and autonomous driving. To facilitate DSA designs, programmers use high-level synthesis (HLS) to compile a high-level description written in C/C++ into a design with low-level hardware description languages that eventually synthesize DSAs on circuits. However, creating a high-quality HLS design still demands significant domain knowledge, particularly in microarchitecture decisions expressed as \textit{pragmas}. Thus, it is desirable to automate such decisions with the help of machine learning for predicting the quality of HLS designs, requiring a deeper understanding of the program that consists of original code and pragmas. Naturally, these programs can be considered as sequence data. In addition, these programs can be compiled and converted into a control data flow graph (CDFG). But existing works either fail to leverage both modalities or combine the two in shallow or coarse ways. We propose \model, a model that allows interaction between the source code sequence modality and the graph modality 
in a deep and fine-grained way. 
To alleviate the scarcity of labeled designs, a pre-training method is proposed based on a suite of compiler's data flow analysis tasks. 
Experimental results show that \model reduces the RMSE of design performance predictions by up to $22\%$, and identifies designs with an average of $1.10\times$ and $1.26\times$ (up to $8.17\times$ and $13.31\times$) performance improvement in design space exploration (DSE) task compared to HARP and AutoDSE, respectively.

\end{abstract}

\section{Introduction} 
\label{sec-intro}

\nop{


There has been an increasing need for specialized computing systems that can speed up particular applications. As a result, domain-specific accelerators (DSAs) such as application-specific integrated circuits (ASICs) or field-programmable gate arrays (FPGAs) have emerged. 
\nop{
DSAs are developed to enhance performance and energy efficiency by exploiting the characteristics of specific workloads. 
The Tensor Processing Unit (TPU)~\citep{jouppi2017datacenter}, Google's custom-designed ASIC, is a prominent example of a DSA that has been optimized for machine learning workloads and can deliver an orders-of-magnitude faster performance and better energy efficiency than a CPU or GPU.
}
However, designing DSAs is very challenging~\citep{autodse, chi2022democratizing, dse-survey} because they are usually designed using hardware description languages (HDLs) at the register-transfer level (RTL) using Verilog and VHDL, which are only familiar to circuit designers.
To address this challenge, high-level synthesis (HLS)~\citep{cong2011high, cong2022fpga} was introduced.
HLS raises the level of abstraction to C/C++/OpenCL/SystemC, allowing designers to describe a high-level behavioral representation of their design rather than the transition of data in RTL. 

Although HLS tools increase the level of design abstraction, they still require a significant amount of hardware design knowledge expressed through synthesis directives in the form of pragmas. These pragmas specify how computation is parallelized and/or pipelined, how data is cached, how memory buffers are partitioned, etc. (see Section \ref{sec-prelim} for more details).
However, such architecture-specific optimization can usually only be done by hardware programmers and is beyond the reach of the average software programmer. Our objective is to automate and speed up the process of optimizing an integrated circuit (IC) design to make it more accessible for the average software programmer. 
}

Over the past decades, the need for specialized computing systems to accelerate specific applications has grown, leading to the emergence of domain-specific accelerators (DSAs) like application-specific integrated circuits (ASICs) and field-programmable gate arrays (FPGAs). Designing DSAs is challenging because it involves using hardware description languages (HDLs) at the register-transfer level (RTL) with Verilog and VHDL, which are mainly familiar to circuit designers. High-level synthesis (HLS) was introduced to address this by raising the level of abstraction to C/C++/OpenCL/SystemC, allowing designers to describe high-level behavioral representations of their designs. Despite this, HLS tools still require significant hardware design knowledge through synthesis directives in the form of pragmas, which specify computation parallelization, data caching, memory buffer partitioning, etc. These optimizations are typically done by hardware programmers and are beyond the reach of average software programmers. Our objective is to automate and accelerate the optimization of integrated circuit (IC) design, making it more accessible to software programmers.


There is a growing trend to apply machine learning to IC design automation~\citep{huang2021machine}. For example, researchers have developed learning-based methods to predict the quality of HLS designs~\citep{sohrabizadeh2022automated,sohrabizadeh2023robust}, to explore the HLS design space intelligently for optimal resource allocation~\citep{wu2022ironman}, etc. These methods fundamentally rely on an informative representation of an input design for high-quality performance prediction. 
We, therefore, focus on the representation learning for IC designs defined with HLS C/C++ (in short, we call them HLS designs) which are annotated with compiler directives/pragmas. 
Specifically, we aim to design an encoder-decoder framework where the encoder provides powerful representations for the input HLS designs so that the designs' quality can be predicted accurately.

One limitation of the existing representation learning methods for programs and HLS designs is that they usually restrict the model to only using either the source code or the compiler-derived representation, but not both. For example, previous works~\citep{sohrabizadeh2022automated,sohrabizadeh2023robust,wu2022high} compile the HLS code into LLVM intermediate representation, which is then further transformed into a graph representation before a graph neural network (GNN) is used to encode it.
Meanwhile, \citep{wang2021codet5,cubert,feng2020codebert,guo2020graphcodebert} directly apply a large language model (LLM) to the source code to obtain the representations that catch the semantics of general computer programs. 

However, we argue that only utilizing either one of the modalities is not good enough to obtain a comprehensive program representation. One the one hand, the graph modality tends to ignore the semantic information in the source code which is helpful to understand a program's behavior. For example, in CDFG, it is difficult for GNN to understand the functionality of a call site, particularly to ones such as standard libraries (e.g., \texttt{glibc}). What is worse, a statement such as ``\texttt{A[i][j] *= beta;}'' would be converted to a relatively large and complex subgraph in the CDFG making it difficult for the model to understand the semantic meaning.
On the other hand, two source code programs with similar semantics and functionalities could have significantly different latency and communication requirements. This is where the lower-level control-flow structure of the programs can help. 
Therefore, a novel model that effectively utilize information from both modalities could be the key to generating powerful representations of HLS designs and general programs.  
In this paper, we propose \model (\emph{\underline{Prog}}ram representation learning combining the source \emph{\underline{S}}equence and the control data flow \emph{\underline{G}}raph) for a unified representation learning  that leverages both the source code modality and an enriched CDFG graph modality, with pre-training performed on both modalities. 
To handle the interaction between source code and CDFG graph, we propose two innovative designs in the architecture: (1) An attention-summary architecture for coarse interaction between the two modalities; (2) A fine-grained node-to-token message passing mechanism to enable further collaboration between the two modalities. We also propose a novel pre-training method based on predicting node-node relationships for compiler analysis tasks which helps the GNN encoder to address the label scarcity issue. Experiment results show the proposed \model achieves a state-of-art performance on design quality prediction and design space exploration.

\nop{
Our contributions in this paper can be summarized as follows:
\begin{itemize}
    \item This work is among the first to tackle the emerging problem of HLS performance prediction and design automation with the recently popular transformer model.
    \item We propose a novel neural network architecture based on multi-modality learning, with graph summary augmentation and node-token alignment to facilitate collaboration between the two modalities efficiently and effectively. 
    \item We propose a new pretraining method for GNNs tailored for computer programs via CDFGs. To the best of our knowledge, we are among the first to explore pre-trained GNNs for assembly code derived CDFGs.
    \item Experiment results show the proposed \model achieves a state-of-art performance on design quality prediction and design space exploration.
\end{itemize}

The rest of the paper is organized as follows: Section~\ref{sec-prelim} introduces the design quality prediction task; Section \ref{sec-related} surveys existing works on related areas; Section~\ref{sec-model} presents \model; Section~\ref{sec-exp} illustrates experimental results; Section~\ref{sec-conc} gives a conclusion to this work and discusses on future possibilities. Our code is available online\footnote{\url{https://anonymous.4open.science/r/software-gnn-DF81/}}.
}
\section{Preliminaries}
\label{sec-prelim}


\subsection{HLS Design and Optimization Pragmas}
\label{subsec-probdef}

The goal of this paper is to train a model to effectively predicts the quality of the HLS design, which 
is a C/C++ program 
with inserted pragmas serving as design specification.
The quality of a design is  measured by its latency in cycle counts (perf), the utilization rate of block RAM (util-BRAM), digital signal processors (util-DSP), flip-flop (util-FF), and lookup-tables (util-LUT)~\citep{autodse,sohrabizadeh2022automated}. 


We specifically consider the optimization pragmas of the Merlin Compiler, an open-source tool widely used for HLS designs\footnote{\url{https://github.com/Xilinx/merlin-compiler}}. The Merlin Compiler provides three types of optimization pragmas, namely \texttt{PIPELINE}, \texttt{PARALLEL}, and \texttt{TILE} to define the desired microarchitecture~\citep{autodse}. 
As shown in Code~\ref{code:mvt} in the Appendix, 
these pragmas can be applied at the loop level and offer control over the type of pipelining, the parallelization factor, and the amount of data caching. 
\nop{
With carefully selected pragmas and their corresponding parameter values,
the resulting accelerator can be $10\times$ faster than a single-core CPU. However,
without any pragma insertion, the resulting hardware is $13\times$ slower than a CPU. 
}
Table \ref{tab:pragma} summarizes the parameter space of these pragmas. 
For a given program $P$, any change in the option of any of the pragmas 
results in a different design $D$ with a unique microarchitecture. For example, the ``fg'' option in pipelining refers to the case where all the inner loops are unrolled (parallelized with separate logic) and each parallel unit is pipelined. The ``cg'' option, on the other hand, results in coarse-grained processing elements (PEs) that are pipelined together. For example, it can create pipelined load-compute-store units. The \texttt{PARALLEL} and \texttt{TILE} pragmas take numeric values that determine the degree of parallelization and loop tiling, respectively.
\begin{table}[h]
    \centering
    \caption{Target pragmas with their options.}
    \label{tab:pragma}
    \begin{tabular}{c|c|c}
    \hline
       Pragma  & Parameter Name & Parameter Space \\
       \hline
       \texttt{{PARALLEL}}  & factor & integer \\
       \texttt{PIPELINE} & mode & ``cg'', ``fg'', off\\
       \texttt{TILE} & factor & integer \\
       \hline
    \end{tabular}
\end{table}

\subsection{Hierarchical Graph Representation of HLS Designs~\label{sec:harp-graph}}

\begin{figure}
    \centering
    \includegraphics[width=0.6\linewidth]{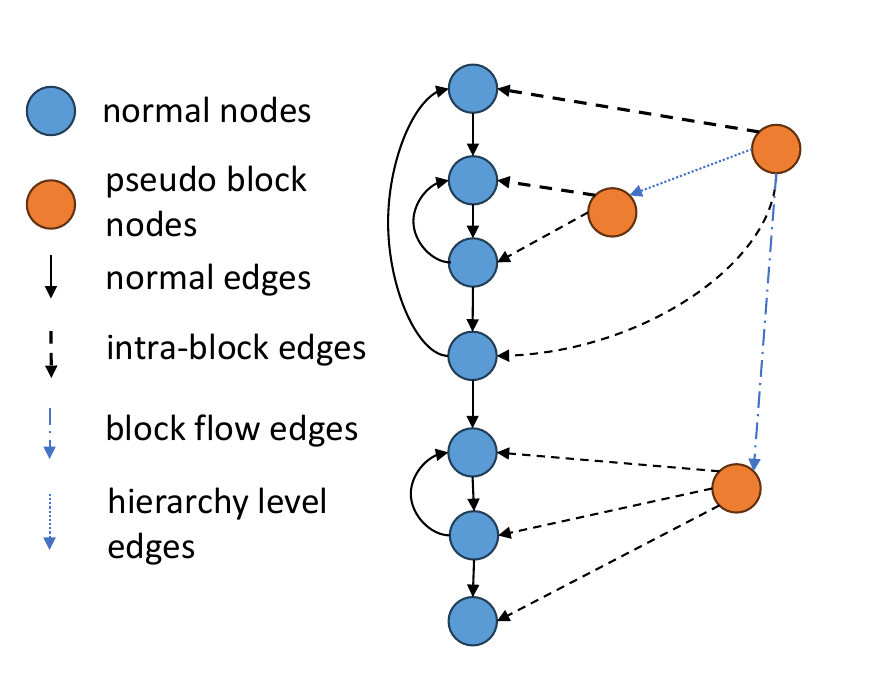}
    \caption{An Illustration of HARP control data flow graph. Compared with a normal CDFG, it has additional block nodes and three types of edges: intra-block edges, block-flow edges, and hierarchy-level edges.
    }
    \label{fig:harp}
\end{figure}

We leverage HARP's approach~\citep{sohrabizadeh2023robust} to generate the hierarchical graph representation of an HLS design, which is an enriched CDFG with extra block nodes and their connections. 
Figure \ref{fig:harp} depicts an illustration of a \textit{HARP graph}. Specifically, given the source code $C=(c_1,\ldots,c_I)$ ($c_i$, $i=1,\ldots,I$ denotes the $i$-th token of the source code), 
it is first transformed into an LLVM~\citep{lattner2004llvm} intermediate representation (IR), and further converted into a CDFG\footnote{Strictly speaking, it is a modified \programl graph with additional call relations between instructions and explicit nodes for operands with additional pragma nodes, but for convenience and without loss of generality, we use the term ``CDFG'' in this paper.}. Then to insert hierarchical information into the graph, auxiliary nodes are added into the graph where each auxiliary node represents a distinct LLVM IR block. Each of these blocks is a sequence of instructions that has a single entry point and a single exit point. Each auxiliary node has three types of edges: the edges to all instruction and data nodes within that block (intra-block edges), the edges to the previous and next block (block flow edges), and the edges building connections based on the hierarchy level of the "for" loops in the C/C++ code (hierarchy level edges). HARP~\citep{sohrabizadeh2023robust} shows that the hierarchical graph representation helps propagate the long-range dependency information in the graph, which helps it learn a better graph representation.

\nop{
\textbf{Task Definition} \enspace The model $f(D_i)$ predicts $\hat{\bm{y}_i}$ for a given input design $D_i=(P_i,\zeta_i)$. The model is trained on a set of labeled designs $ \mathcal{D}^{(\mathrm{train})} = \{ (D_i, \bm{y}_i) \}_{i=1}^{N}$ coming from a variety of programs with their pragmas\footnote{Due to the combinatotrial nature of design space, we follow \citep{sohrabizadeh2022automated} to exclude invalid designs from the training and testing set.}.
}

\nop{
\textbf{Source Code and CDFG} \enspace In this paper, we follow GNN-DSE's~\citep{sohrabizadeh2022automated} approach
to compile the source code $C=(c_1,\ldots,c_I)$ where $I$ denotes the sequence length via LLVM~\citep{lattner2004llvm}, and transform the assembly code further into a CDFG\footnote{Strictly speaking, it is a \programl graph with additional call relations between instructions and operands and with additional pragma nodes, but for convenience and without loss of generality, we use the term ``CDFG'' in this paper.} denoted as $G$. $G=(V,E,l_V,l_E)$ where $V$ denotes the node set, $E$ denotes the edge set, and $l_V$ ($l_E$) denotes the node (edge) labeling function that maps each node (edge) into a list of node (edge) attributes. The pragmas $\zeta$ are placed in the source code and can be transformed into the CDFG as nodes. Therefore, a design $D=(P,\zeta)$ can also be represented as a source code sequence $C$ and a CDFG $G$, i.e., $D=(C,G)$. 
}

\nop{
\textbf{Cross-Modality Alignment} \enspace  
The LLVM compiler outputs the line and column numbers associated with some of the assembly instructions, which we use to construct a one-to-many alignment $M$ mapping $v_k \in V$ to a set of source code tokens on that line $\{c_j\}$, i.e. $M(v_k)=\{ c_j \}$. The reverse mapping $M^{-1}$ maps a token $c_j$ to nodes $\{ v_k \}$ such that $c_j \in M(v_k)$. More details can be found in the supplementary material. 
 }

\section{Proposed Method: \model}
\label{sec-model}

In this section, we first describe the overall encoder-decoder architecture of \model. Then, we focus on our novel encoder with a graph summary augmented sequence representation, and a fine-grained node-to-token alignment for the unification of the two modalities. Finally, we introduce a novel pre-training framework for program graphs.

\subsection{Overall Architecture for Design Quality Prediction} 
\label{subsec-overal-nn}

\begin{figure*}
\centering
\includegraphics[width=0.7\textwidth]{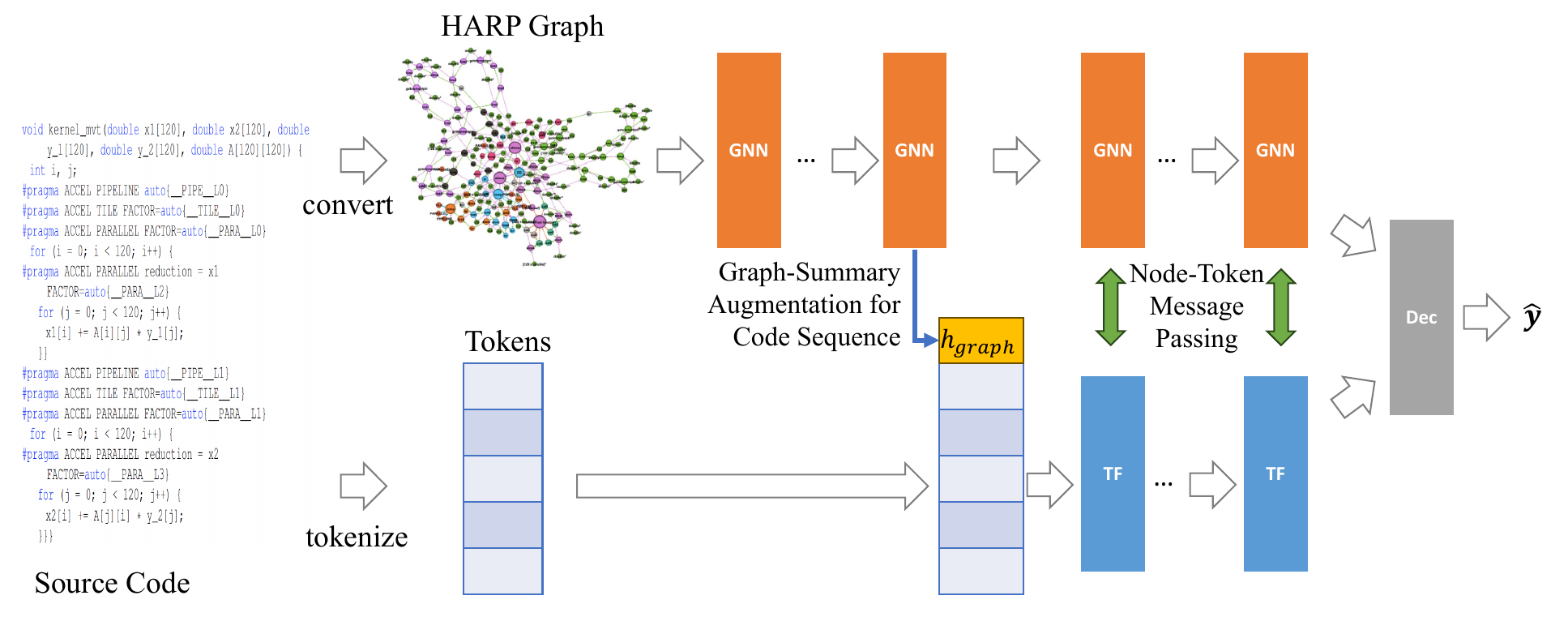}
\caption{The overall diagrams of \model. ``GNN'', ``TF'', and ``Dec'' refer to Graph Neural Network Layer, Transformer Layer, and Decoder, respectively.  
}
\label{fig:model}
\vspace*{-4mm}
\end{figure*}

Given a design $D$ with source code $C$ and HARP graph $G$, the overall model $f(D)=f(C,G)$ first encodes designs 
into a set of embeddings, and then generates predictions $\hat{\bm{y}}$ with a multilayer perceptron (MLP) based decoder. Figure~\ref{fig:model} depicts the overall diagrams of our model. Let $\bm{y}$ indicate the ground-truth targets (i.e., perf, util-BRAM, util-DSP, util-FF, and util-LUT). 
Our objective is to minimize the loss function that measures the mean squared error (MSE) between $\bm{y}$ and $\hat{\bm{y}}$, i.e., $\mathcal{L}_{\mathrm{task}}=|| \hat{\bm{y}}-\bm{y} || ^2$. 

Since one modality is the source code sequence, and the other is the HARP graph, it is natural to adopt a transformer model on $C$ and a GNN model on $G$, which produce token representations $\{ \bm{h}_j \in \mathbb{R}^{d} | j \in  \{ 1,\ldots,I \} \}$ via the transformer's self-attention mechanism, and node representations $\{ \bm{h}_k \in \mathbb{R}^{d} | k \in  \{ 1,\ldots,|V| \} \}$ via the message passing mechanism, respectively. $d$ denotes the embedding dimension. The starting token $c_1$'s embedding is then taken as the source code summary, $\bm{h}_{\mathrm{src}} \in \mathbb{R}^{d}$, and a graph-level aggregation can be performed on the node embeddings serving as the graph summary, 
$\bm{h}_{\mathrm{graph}} \in \mathbb{R}^{d}$. The encoder outputs the concatenation of the two modalities summaries, $\mathrm{concat} (\bm{h}_{\mathrm{src}}, \bm{h}_{\mathrm{graph}} )$, and lets the MLP-based decoder generate predictions.

This model serves as the foundation of our architecture. However, it solely relies on the MLP-based decoder to manage the interaction between the two modalities. We denote this simplified version of our model as \simpleConcat.  


\subsection{\sumInt: Graph-Summary-Augmented Sequence Representation}
\label{subsec-gsa-sr}

One limitation of the \simpleConcat encoder is the shallow and ineffective modeling of the interaction between $C$ and $G$. We propose a novel yet simple way to address the issue, by making the following observation: The transformer operates on the sequence of tokens $C=(c_1,\ldots,c_I)$ by enabling every token to pay attention to every other token. That is,
\nop{
, and thus the embedding of the starting token $c_1$ which is initialized as a special summary token such as ``\texttt{[cls]}'', can be treated as the representation of the entire $C$. Mathematically, we can formulate 
}
\begin{equation}
\label{eq-transformer}
\bm{h}_{\mathrm{src}} = AGG(g_{\mathrm{att}} \big(  \bm{h}^{(0)}_{c_1},\ldots,\bm{h}^{(0)}_{c_I} \big))
\end{equation}
where $AGG$ can be any aggregation function, and $g_{\mathrm{att}}$ denotes the multi-layer self-attention encoder of a transformer model, capturing the interaction between pairwise source code tokens, $\bm{h}^{(0)}_{c_j}$ stands for the $j$-th token's initial embedding\footnote{This is usually implemented by looking it up in a dictionary that maps each token ID into a $d$-dimensional learnable vector representing the initial embeddings.}, $\bm{h}_{\mathrm{src}}$ denotes the final program-level source code embedding.

Based on the above observation, we propose to insert the graph summary $\bm{h}_{\mathrm{graph}}$ to the beginning of the sequence, forming an augmented sequence representation $C^{\mathrm{(aug)}} = ( \bm{h}_{\mathrm{graph}},c_1,\ldots,c_I)$ as input to the transformer\footnote{This is equivalent to augmenting the initial embedding lookup dictionary with a special token initialized as the output of a GNN.}. 
Overall, 
\begin{equation}
\label{eq-sumInt}
\bm{h}_{\mathrm{src}} =  AGG(g_{\mathrm{att}} \big(  \bm{h}_{\mathrm{graph}}, \bm{h}^{(0)}_{c_1},\ldots,\bm{h}^{(0)}_{c_I} \big)).
\end{equation}

We name such an encoder as \sumInt (\emph{\underline{S}}ummary \emph{\underline{I}}nteraction), since it first performs GNN with $L_1$ layers on $G$ to obtain a summary, and let the expressive transformer of $L_2$ layers handle the pairwise attention between tokens and that summary embedding, which efficiently allows cross-modality interaction. In other words, the graph is treated as a derivative of the source code whose summary embedding is used to augment the source code sequence. During training, the gradients back-propagate through $\bm{h}_{\mathrm{graph}}$ to the GNN, updating both the GNN and the transformer.

\nop{It is noteworthy that we are not among the first to propose such a concatenation-based method for interaction between a graph modality and a sequence modality. \greaseLM~\citep{zhang2022greaselm}, for example, updates the summary of a graph and the summary of a sequence first, i.e., $\mathrm{concat}(\bm{h}_{c_{cls}}^{(l+1)}, \bm{h}_{\mathrm{graph}}^{(l+1)}) =  \mathrm{MLP}\big(\mathrm{concat} ( \bm{h}_{c_{cls}}^{(l)}, \bm{h}_{\mathrm{graph}}^{(l)} ) \big)$, where $c_{cls}$ is a special token. And then it applies GNN and transformer again to the original graph and sequence. However, we note that our \sumInt explicitly appends the summary embedding from the other modality as another token, which does not require any modifications to the architecture of language models. 
}

\subsection{Full Model \model: Leveraging Fine-grained Node Token Interaction}

While \sumInt enables interaction between the graph and tokens, the graph-level summary is too coarse for the model to fully exploit the information from both modalities. 
Intuitively, the information exchange between two modalities would be more effective if the interaction happens in node/token level. A straightforward way is to utilize a cross attention module to all node embeddings $\bm{h}_{v_1}, \ldots,\bm{h}_{v_{|V|}}$ and token embeddings $\bm{h}_{c_1},\ldots,\bm{h}_{c_I}$. However, since there could be thousands of nodes and tokens for an HLS design, the computation overhead is too expensive.
So a more efficient way to leverage fine-grained node token interactions is needed.

\begin{figure}
    \centering
    \includegraphics[width=0.8\linewidth]{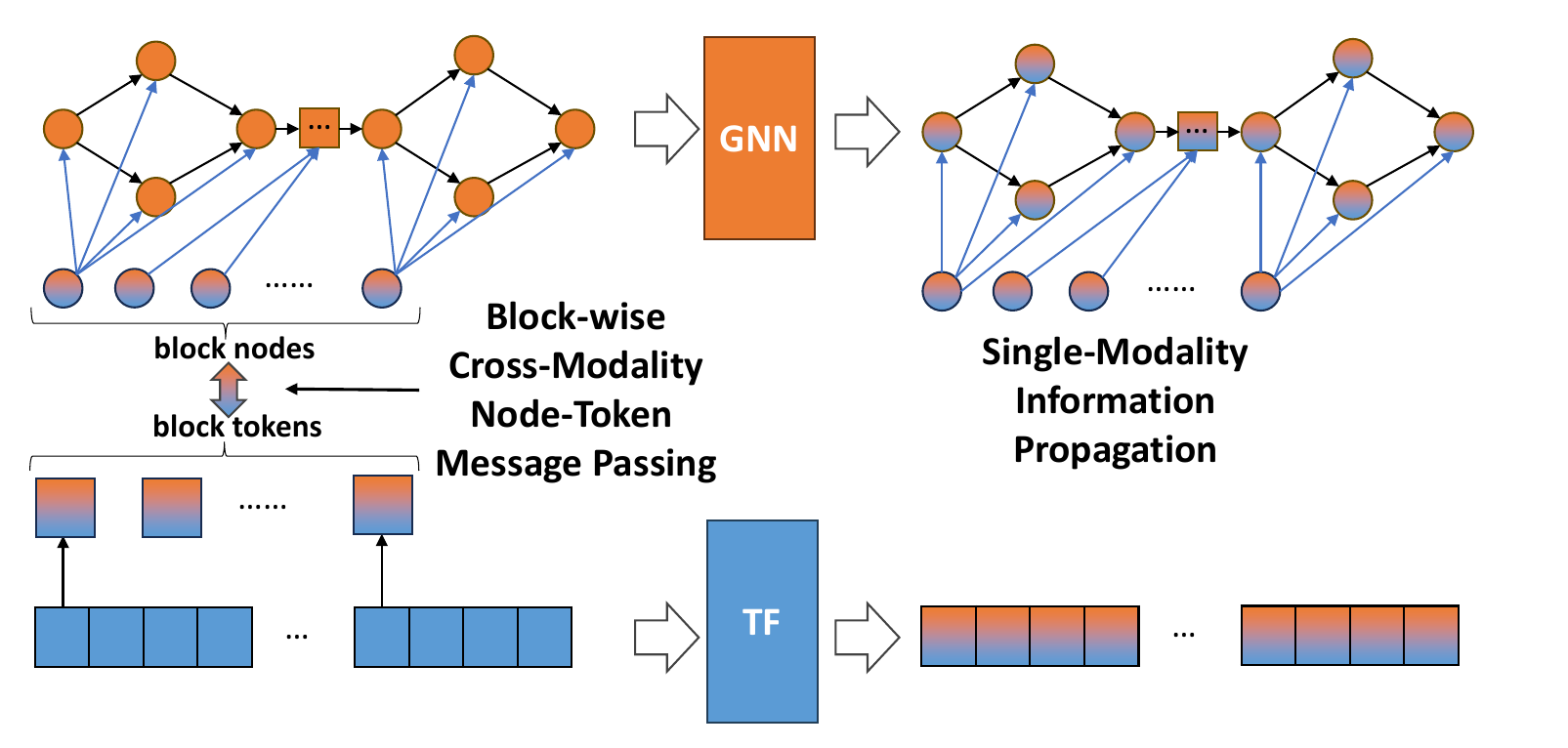}
    \caption{Illustration of the node-token message passing mechanism. The cross-modality information is first exchanged via block nodes and block tokens. Then the information is propagated to normal nodes and tokens through the GNN and transformer layers, respectively.
    }
    \label{fig:node_token}
\end{figure}

Recall that there are auxiliary nodes in the HARP graph that stand for the LLVM-IR blocks (see Sec \ref{sec:harp-graph} for more details). Meanwhile, the source code is segmented into multiple chunks so that the length of each chunk is within the input length limit of the transformer. 
Let $\bm{h}_{v_{a_1}},\ldots,\bm{h}_{v_{a_N}}$ denote the embeddings of auxiliary block nodes and let $\bm{h}_{c_{s_1}},\ldots,\bm{h}_{c_{s_M}}$ indicate the embeddings of the summary tokens in source code chunks.  
Since the auxiliary block nodes in the graph modality and the chunks of source codes provide an intermediate granularity between graph/program and node/token level, we propose to utilize them to conduct a hierarchical node/token interaction, which is illustrated in Figure \ref{fig:node_token}. 
The information between two modalities are first exchanged between the block nodes and the summary tokens via the following cross-modality message passing mechanism inspired by message passing GNNs:
\begin{equation}
\label{eq:block-wise-interaction}
    \begin{array}{rl}
\bm{h}^{\prime}_{v_{a_k}} &= \bm{h}_{v_{a_k}} +
\mathrm{MLP}_2 \Big( \sum_{j} \alpha_{k,j} \mathrm{MLP}_1 (\bm{h}_{c_{s_j}}) \Big),
\\
\bm{h}^{\prime}_{c_{s_j}} &= \bm{h}_{c_{s_j}} +
\mathrm{MLP}_4 \Big( \sum_{k} \alpha_{j,k} \mathrm{MLP}_3 (\bm{h}_{v_{a_k}}) \Big),
\end{array}
\end{equation}
where the attention coefficients are computed via a dot product attention with learnable weight matrices $\bm{W}_1 \in \mathbb{R}^{d \times d}$ and $\bm{W}_2 \in \mathbb{R}^{d \times d}$, $\alpha_{k,j} = \textrm{Softmax} \left(
\frac{(\bm{W}_1\bm{h}_{v_k})^{\top} (\bm{W}_2\bm{h}_{c_j})}
{\sqrt{d}} \right).$

Then, the exchanged information is propagated to each node and token via a GNN and a transformer layer, respectively. Specifically, for a node $v_i$ (token $c_i$) that is not a block node (summary token), let $\bm{h}^{\prime}_{v_i}=\bm{h}_{v_i}$ ($\bm{h}^{\prime}_{c_i}=\bm{h}_{c_i}$), the second step can be written as 
\begin{equation}
\begin{array}{cc}
    &\bm{h}^{\prime\prime}_{v_i} = \bm{h}^{\prime}_{v_i} +
\mathrm{MLP}_6 \Big( \sum_{j} \alpha_{i,j} \mathrm{MLP}_5 (\bm{h}^{\prime}_{v_j}) \Big) \\
    &\bm{h}^{\prime\prime}_{c_1},\ldots,\bm{h}^{\prime\prime}_{c_I} = Attention(\bm{h}^{\prime}_{c_1},\ldots,\bm{h}^{\prime}_{c_I})
\end{array}
\end{equation}

Such cross-modality interaction enables \textbf{\textit{fine-grained}} interaction between the two modalities so that more informative embeddings for the final prediction task can be generated. As an additional benefit, the interaction step is significantly more efficient than the full cross-attention because the number of auxiliary nodes and summary tokens is usually small.
To allow \textbf{\textit{deep}} cross-modality interaction, we perform the above node-token message passing $L_2$ times where $L_2$ is the number of transformer layers, e.g., 6 for the pre-trained \codetfive model used in our experiments. In each of the $L_2$ layers, \model performs the self-attention encoder on $C^{(aug)}$, and executes GNN on $G$, followed by the node-token interaction. 

\subsection{Pretraining GNNs for Graph Modality}


Generating ground-truth targets with an HLS simulator is slow, resulting in a scarcity of labeled data. To mitigate this issue, we propose utilizing pre-training tasks. While there is extensive work on pre-training transformer models with code~\citep{wang2021codet5,feng2020codebert,cubert}, our focus is on pre-training GNNs for graph modality. Existing self-supervised tasks for GNNs are for general graphs instead of CDFG; thus, we propose employing data flow analyses as self-supervised learning tasks. Data flow analysis is fundamental to modern compiler technology~\citep{cummins2021programl} and necessitates that GNNs extract crucial information from a program's structure. Furthermore, these tasks can be effectively addressed by non-ML techniques, allowing us to easily obtain a substantial set of labeled data for pre-training.

In particular, we select four data analyses tasks: (1) reachability: if a node can be reached from another node, (2) dominators: if every control-flow path to an instruction node passes through another node, (3) data dependencies: if a variable is defined in an instruction and used in another instruction, and (4) liveness: if a variable is live-out of a statement $n$. More detailed definitions of these tasks can be found in~\citep{cummins2021programl}. 
These tasks cover a full range of forward and backward analyses, and control and data analyses. In addition, these tasks focus on predicting the relationship between two nodes in a CDFG. Such node-level tasks help the GNN to learn meaningful node embeddings, which is the foundation of generating good graph embeddings. 
Each task can be viewed as a binary classification problem. Given a pair of nodes $v_i, v_j$ and a label $y_{ij}$ which is a binary label indicating if the nodes have a particular relationship, we employ the cross entropy loss for pre-training loss.
\nop{
\begin{equation}
\begin{split}
    \mathcal{L}_{\mathrm{CE}} = &-y_{ij}\log(p(v_i,v_j))
    - (1-y_{ij})\log(1-p(v_i,v_j))
\end{split}
\end{equation}
}
\nop{binary focal loss~\citep{lin2017focal} to optimize the parameters.
\begin{equation}
\begin{split}
    \mathcal{L}_{\mathrm{focal}} = &-y_{ij}(1-p(v_i,v_j))^\beta \log(p(v_i,v_j))\\
    & - (1-y_{ij})(p(v_i,v_j))^\beta\log(1-p(v_i,v_j))
\end{split}
\end{equation}
where $p(v_i,v_j)$ is the predicted probability that $y_{ij}=1$ and $\beta\ge0$ is a hyper-parameter.
Focal loss is a good replacement for cross-entropy loss when the numbers of positive and negative samples are imbalanced. The modulated term helps the model to focus on hard misclassified examples.
}

Normally after pre-training, we would directly fine-tune the pre-trained GNN for the downstream task. However, the pre-training dataset does not contain any pragma nodes, which is important for predicting the quality of the HLS design. Therefore, we propose to use the pre-trained node embeddings as guidance to train a new (target) GNN for the downstream task. 
Specifically, given a graph with pragma nodes, denoted as $G$, we would generate a corresponding graph without pragma nodes, designated as $G'$. Then for a node $v$ that appears in both $G$ and $G'$, we would compute its embedding in $G'$ with the pre-trained GNN and compute its embedding in $G$ with the GNN to be trained. Then, we would maximize the cosine similarity between the two embeddings with the following loss $\mathcal{L}_{\mathrm{guide}}=1-\cos\langle g_{\mathrm{cont}}(\bm{h}_{v,G}), \bm{h}_{v,G'}\rangle$ where $g_{\mathrm{cont}}$ is a continuous function (e.g., MLP, identity function).
In this way, the target GNN would learn how to extract useful node-level information from the pre-trained GNN, which would in turn improve the quality of graph-level embeddings.



\section{Experiments}
\label{sec-exp}


Here we present the main experiment results. Additional experiments are provided in the Appendix.

\subsection{Model Hyperparameters and Training Details}

During training, we combine the proposed loss functions including $\mathcal{L}_{\mathrm{total}} = \mathcal{L}_{\mathrm{task}} + \gamma_1 \mathcal{L}_{\mathrm{fineAlign}} + \gamma_2 \mathcal{L}_{\mathrm{coarseAlign}} + \gamma_3 \mathcal{L}_{\mathrm{guide}}$, where $\gamma$s are hyperparameters controlling the weight for the different loss terms. During inference, we apply the encoder-decoder architecture to obtain $\hat{\bm{Y}}$.


We set the maximum number of tokens to 64 for the tokenizer, and chunk each source code into multiple subsequences to handle the long input source code sequence. We leave the exploration using more advanced modeling for long sequences such as \cite{xiao2023efficient} as future work. Since the task is on the whole program level, for each subsequence, we use the final embedding of the initial token (``\texttt{[cls]}'') as the summary of each subsequence (for \sumInt and \model, an additional MLP is applied to project $\bm{h}_{\mathrm{src}} = \bm{H}_{\mathrm{src}}[0:1]$ from dimension 1024 to 512), and aggregate all summaries into a final sequence-level embedding (denoted as $\bm{h}_{\mathrm{src}}$ in the main paper) which is fed into the decoder. For the two-modality models, the decoder receives the concatenation of $\bm{h}_{\mathrm{src}}$ and $\bm{h}_{\mathrm{CDFG}}$ as described in the main paper. 

The decoder consists of 6 sequentially stacked layers that project the input to a scalar. If the model is of a single modality, the MLP decoder has hidden dimensions 512-256-128-64-32-16-1. If the model is of two modalities, the MLP decoder has hidden dimensions 1024-768-512-256-128-61-1. The above scheme is administered consistently to all the methods for a fair comparison. Since we have 5 target metrics to predict as mentioned in Section 4.1 of the main paper, we use 5 MLPs applied on the input embeddings to transform them into the final $\hat{\bm{y}} \in \mathbb{R}^{5}$. We use the Exponential Linear Unit (ELU) function~\cite{clevert2015fast}.

Our framework is implemented with PyTorch, PyTorch Geometric, Transformers, etc\footnote{We will release our code and data upon acceptance.}. Training is performed on a server with NVIDIA Tesla V100 GPUs. We employ the AdamW optimizer~\citep{loshchilov2017decoupled} with the initial learning rate tuned for each model using a validation set. We perform training 
with $\gamma_1=\gamma_2=\gamma_3=1$ 
over 1000 epochs with the best model selected based on a validation set for final adaptation and testing. 

For the pre-trained GNN, 
we use utilize GNN with 5 transformer convolutional layers~\citep{shi2020masked} as encoders and a 2-layer MLP as the decoder for each data analysis task. 
We use a training set with 276,197 graphs. 
The $\beta$ in focal loss is set to 2.
We employ a validation set with 500 graphs to select the best pre-trained GNN. The $\beta$ in focal loss is set to 2.

\subsection{Dataset and Evaluation Protocol}
For the purpose of this study, we assembled a database of medium-complexity kernels that function as fundamental building blocks for larger applications. We selected a total of 42 kernels from two well-known benchmark suites, namely, the MachSuite benchmark~\citep{machsuite} and the Polyhedral benchmark suite (Polybench)~\citep{polybench}. 
The kernels in the database were chosen to have a broad range of computation intensities, including linear algebra operations on matrices and vectors (e.g., BLAS kernels), data mining kernels (e.g., \corr and \cov), stencil operations, encryption,
and a dynamic programming application.

The database is a new version of datasets released in~\cite{bai2023towards}, generated by the AMD/Xilinx HLS tool version 2021 to implement the design, with the AMD/Xilinx Alveo U200 as the target FPGA and a working frequency of 250MHz. For each kernel, we perform a random split with the training, validation, and testing ratio being 70:15:15.
For each design point, we recorded the \texttt{latency} in terms of cycle counts, as well as the resource utilization for \texttt{DSP}, \texttt{BRAM}, \texttt{LUT}, and \texttt{FF}. These targets are normalized following the same procedure in~\cite{bai2023towards,sohrabizadeh2022automated}.
The statistics of the dataset are presented in Table~\ref{table-dataset}. 
The dataset will be available upon paper acceptance.


\subsection{Model Setup and Hyperparameters}

We follow \citep{sohrabizadeh2023robust} to generate the HARP graphs. We adopt $L_1=8$ layers of \tfconv~\citep{shi2020masked} with a jumping knowledge network~\citep{xu2018representation} as the final node embedding aggregation method. The embedding dimension $d=512$. For the source code, we use \codetfive~\citep{wang2021codet5} with $L_2=6$ layers to embed the source code\footnote{Specifically, we use \textsc{CodeT5-small} from \url{https://huggingface.co/Salesforce/codet5-small} to initialize the transformer encoder for source code, and fine-tune the whole model.}. AutoDSE defines a variable for each pragma, as shown in Code~\ref{code:hls}, that is a placeholder for the option of the pragma. 
Since the pragmas $\zeta$ must be reflected in the input source code, for each design, we add the pragma options to their respective variables, e.g., we change ``\texttt{\_\_PARA\_\_L0\_\_}'' to ``\texttt{\_\_PARA\_\_L0=1}'', ``\texttt{\_\_PIPE\_\_L2}'' to ``\texttt{\_\_PIPE\_\_L2=flatten}'', etc. 
We set the maximum number of tokens to 64 for the tokenizer, and chunk each source code into multiple subsequences to handle the long input source code sequence. The summaries of all subsequences are aggregated into the final representation for the decoder. We report the full hyperparameters in the appendix.

\begin{table}
  \begin{center}
    \caption{Dataset statistics. ``\textbf{\#D}'', ``\textbf{\#P}'', ``\textbf{A\#P}'', ``\textbf{A\#T}'', ``\textbf{A\#N}'', ``\textbf{A\#E}'', and ``\textbf{A\#MP}'' denote ``\# designs'', ``\# programs'', ``avg \# pragmas per design'', ``avg \# tokens per program'', ``avg \# nodes per program's CDFG'', and ``avg \# edges per program's CDFG'', respectively.
    }
    \begin{tabular}{l|llllll}
    \label{table-dataset}
    \textbf{Dataset} &
    \textbf{\#D} &
    \textbf{\#P} & \textbf{A\#P} &
    \textbf{A\#T} & \textbf{A\#N} & \textbf{A\#E}\\
      \hline
    Vitis 2021     & 10,868 & 40 & 8.1 & 1286.3 & 354.7 & 1246.4 \\
    \end{tabular}
  \end{center}
\vspace*{-4mm}
\end{table}

\begin{table*}[htbp]
    \centering
    \scriptsize
    \caption{Rooted mean sqaure error (RMSE) of different methods in predicting target values.}
    \label{tab:trans_rmse}
    \begin{tabular}{c|c|cccccc}\toprule
      &   & perf & util-LUT & util-FF & util-DSP & util-BRAM & total \\\midrule
  \multirow{3}{*}{\shortstack{Single\\Modalities\\Model}}   &\codeTwoVec &1.0641 & 0.5462& 0.3103& 0.9989& 0.1555 &3.6150\\
    & \cdfg  & 0.2671	&0.1043	&0.0565	&0.1584	&0.0611	&0.6474\\
      & \codetfive & 0.2077	&0.0985	&0.0619	&0.1881	&0.0597	&0.6159\\\hline
   \multirow{4}{*}{\shortstack{Cross\\Modalities\\Model}}  & \greaseLM & 0.2033& \underline{0.0805}& \underline{0.0499}& \underline{0.1349}& 0.0459&\underline{0.5146}\\
     & \simpleConcat & 0.2181	&0.1232	&0.0532	&0.1381	&\underline{0.0334}	&0.5660\\
     & \sumInt & \underline{0.1591}	&0.1630	&0.0514	&0.1558	&0.0335	&0.5628\\
     & \model & \textbf{0.1481}	&\textbf{0.0709}	&\textbf{0.0406}	&\textbf{0.1084}	&\textbf{0.0242}	&\textbf{0.3923}\\
      \bottomrule
    \end{tabular}
\end{table*}

\subsection{Performance Prediction Results}

We compare the accuracy of performance prediction of \model against three categories of baselines: (1) models of source code modality, \codeTwoVec~\citep{alon2019code2vec} and \codetfive~\citep{wang2021codet5}; (2) models of graph modality, \cdfg~\citep{sohrabizadeh2023robust}; and (3) models of both modalities, \greaseLM~\citep{zhang2022greaselm}. \greaseLM is designed for knowledge graph augmented question-task. It exchanges information from two modalities in the program/graph level, which is too coarse. We also include \simpleConcat, which is a simple concatenation of the summary representations described (Section~\ref{subsec-overal-nn}), and \sumInt which combines the two modalities without fine-grained interaction (Section~\ref{subsec-gsa-sr}). 

Table \ref{tab:trans_rmse} provides a detailed breakdown of the prediction accuracy across different target variables. Notably, our results consistently reveal that the cross-modality model outperforms the single-modality model in terms of rooted mean square error (RMSE). This finding strongly supports our argument for the benefits of integrating multiple modalities within our model architecture.
Furthermore, the comparison between the error rates of \sumInt and \simpleConcat highlights the effectiveness of our graph-summary-augmented sequence representation. 
Moreover, \model surpasses \simpleConcat, \sumInt, and \greaseLM. This outcome underscores the superiority of our fine-grained node token interaction module, enabling more accurate predictions across a diverse range of target variables.
In summary, our experimental results validate the effectiveness of our novel cross-modality program encoder. 

\subsection{Design Space Exploration Results}

\begin{figure}
    \centering
\begin{subfigure}[b]{0.23\textwidth}
    \centering
    \includegraphics[width=\textwidth]{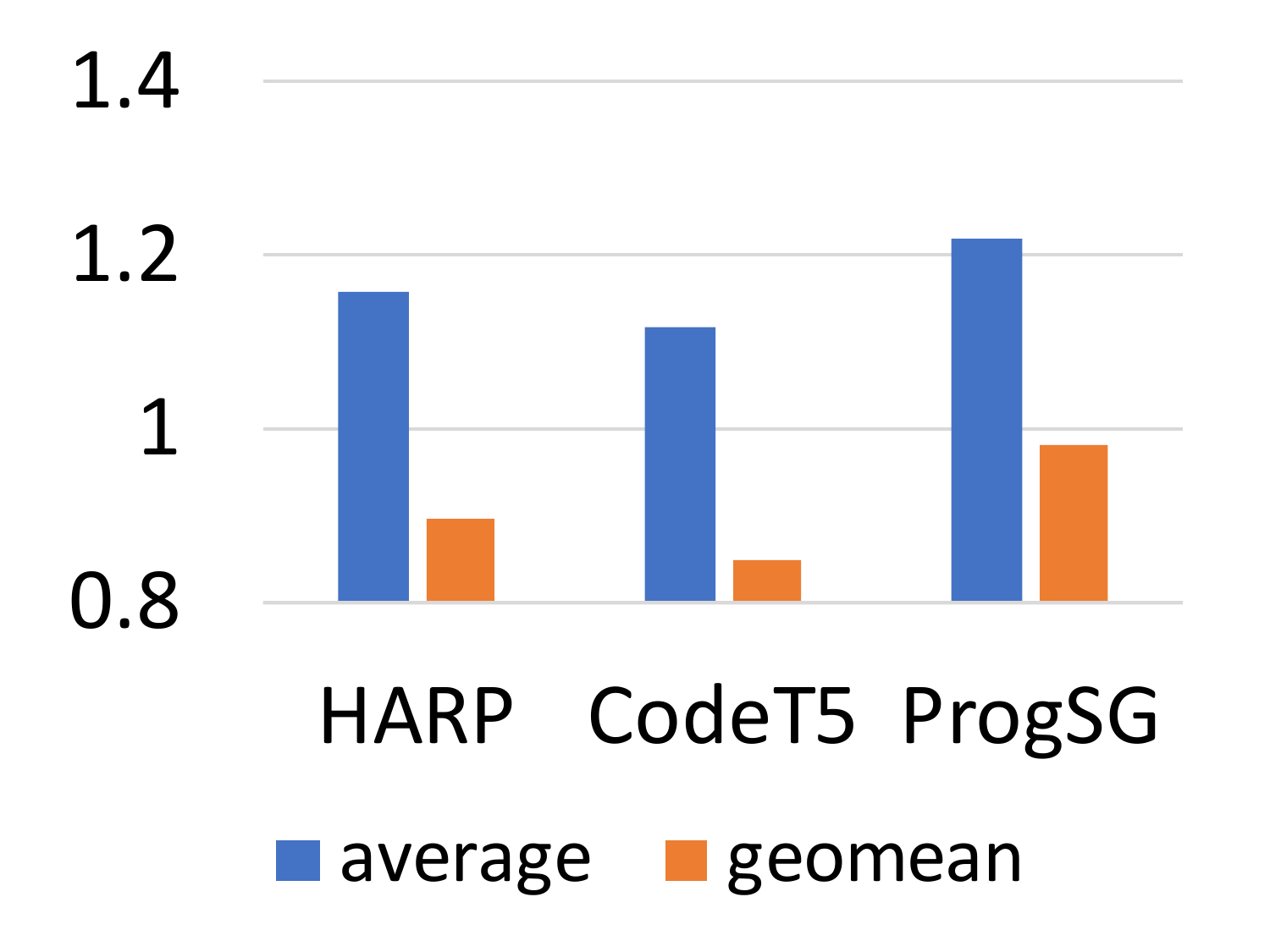}
    \caption{Running each model for one hour.}
    \label{fig:dse_1h}
\end{subfigure}
\hfill
\begin{subfigure}[b]{0.23\textwidth}
    \centering
    \includegraphics[width=\textwidth]{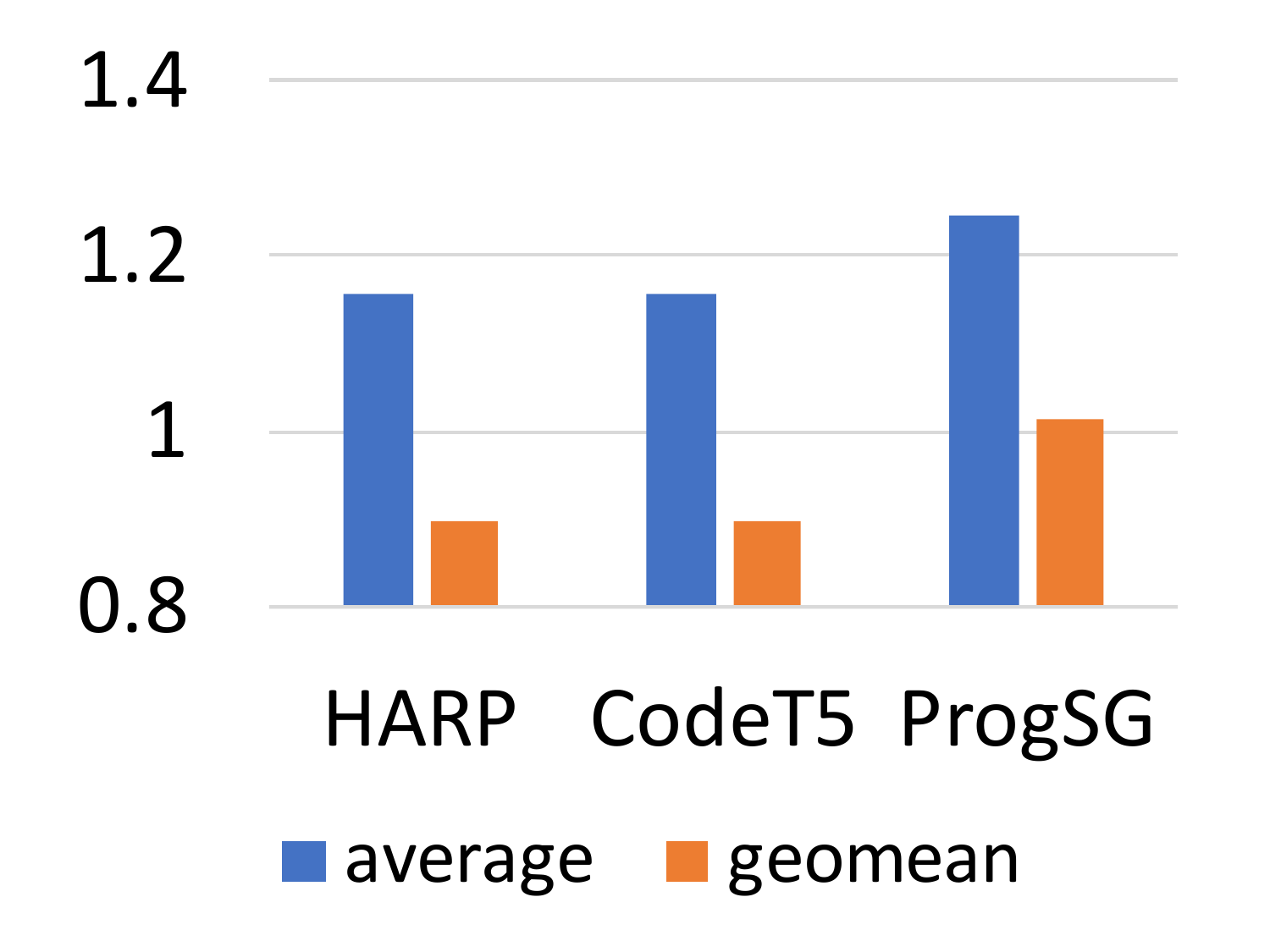}
    \caption{Running each model on 1K candidates returned by HARP. }
    \label{fig:dse_2level}
\end{subfigure}
    \caption{Relative performance improvement of best design found by our model compared to running AutoDSE for twenty-five hours.}
    \label{fig:dse}
\end{figure}

In addition, we evaluate how our method performs in finding the best design of a given kernel, i.e., design space exploration. Following previous studies~\citep{sohrabizadeh2022automated,sohrabizadeh2023robust}, for each kernel we have each model to verify as many designs points as possible in an hour following a heuristic order. The design points with the top 10 predicted performance is recorded. Then we run an HLS simulation to get the ground-truth performance of the selected design points and compare them with the best design point found by running AutoDSE~\citep{autodse} for 25 hours. We use the average speedup between design points found by the model and those found by AutoDSE as the metric to evaluate the performance of each model on DSE task. Figure \ref{fig:dse_1h} shows the average and geomean of the DSE performance of \cdfg, \codetfive, and \model. \model outperforms the two single-modality baselines, revealing that our cross-modality model is superior. In addition, \codetfive is worse than \cdfg in the DSE task, though it has a smaller RMSE in the regression task. We think it is because running \codetfive for inference is slower than running \cdfg, as the GNN is much smaller. As a result, the number of designs verified by \codetfive is smaller. Meanwhile, although \model also suffers from the slow inference, it still manages to find better design points due to its better prediction accuracy.

One way to handle the slow inference speed of \codetfive and \model is to do a two-level design space exploration. That is, \cdfg is first run for an hour to find 1,000 candidate designs with the best predicted performance, then the larger model (\codetfive or \model) is used to select the top 10 designs from them. This two-level approach can simultaneously utilize the efficiency of the GNN model and the effectiveness of larger cross-modality model.
Figure \ref{fig:dse_2level} shows the DSE performance of this two-level approach. It is clear that the performance of \codetfive and \model are significantly improved, showing the advantage of the two-level design space exploration. However, \codetfive still cannot outperform \cdfg, demonstrating that LLM itself might not be powerful enough for our task.

\subsection{Effects of Pre-training}


In addition to our main analysis, we conducted an ablation study to delve deeper into the impact of our Graph Neural Network (GNN) pre-training strategy on model prediction accuracy. The results, as presented in Table \ref{tab:pretrain}, provide insightful observations. Notably, while there is a slight decrease in the accuracy of performance prediction, the prediction accuracy for the other four targets shows a notable improvement ranging from 10\% to 15\%. The drop of accuracy in performance is because we train the model to predict multiple targets simultaneously, and the accuracy of one target might drop while the overall prediction effectiveness improves. Furthermore, when considering the overall prediction accuracy, we observe an improvement of 5.57\%. This substantial boost reaffirms the efficacy of our pre-training approach in refining the model's effectiveness across diverse prediction tasks. 

\begin{table}[]
    \centering
    \caption{Effects of pre-training to the prediction RMSE of our model.}
    \label{tab:pretrain}
    \begin{tabular}{c|ccc}\toprule
      Targets   &  wo pretrain & with pretrain & relative impr.\\\midrule
      perf & 0.1387 & 0.1481 & -6.8\%\\
      util-LUT & 0.0830 & 0.0709 & 14.6\%\\
      util-FF & 0.0461 & 0.0406 & 11.9\%\\
      util-DSP & 0.1084 & 0.1022 & 9.97\%\\
      util-BRAM & 0.0281 & 0.0242 & 13.9\% \\
      \midrule
      total & 0.4163 & 0.3923 & 5.77\%\\
      \bottomrule
    \end{tabular}
\end{table}

\subsection{Attention Visualization}

\begin{figure}
\vspace*{-4mm}
\centering
  \includegraphics[width=0.435\textwidth]{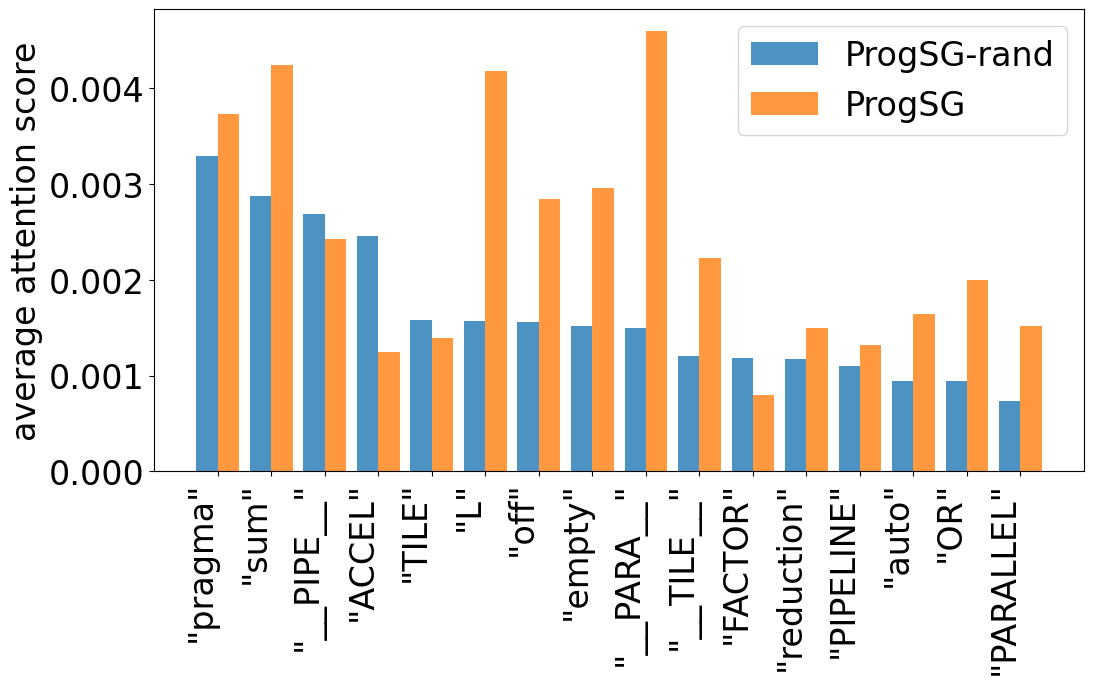}
     \caption{Bar plots of the average attention scores of pragma-related tokens before (\textsc{\model-rand}) and after (\model ) being fine-tuned.}
     \label{fig-att-vis}
\vspace*{-4mm}
\end{figure}

To better understand if the transformer model learns to attend tokens that are relevant to HLS pragma configurations, we visualize the average attention scores of some of the pragma-related tokens in the \gemmNcubed kernel (shown in Code \ref{code:hls}) for the transformer before and after training for our regression task (illustrated in Figure \ref{fig-att-vis}). We can see that 11 out of 15 tokens have higher attention scores after fine-tuning. For the 4 tokens that have lower attention scores (i.e., ``\_\_PIPE\_\_'', ``ACCEL'', ``TILE'', and ``FACTOR''), we can see that they often appear simultaneously with other keywords such as ``PIPELINE'', ``\_\_TILE\_\_'', and ``PARALLEL'', which makes them somewhat redundant. If we compute the summation of their attention scores with the attention scores of tokens that simultaneously appear with them (e.g., ``\_\_PIPE\_\_'' and ``PIPELINE''), we find that the summed attention score increases after training. So the changes in attention score suggest that the transformer model does learn to attend to the pragma-related tokens, which are important to predicting the quality of an HLS pragma configuration, even though these tokens are not included in its pre-training stage.


\vspace{3mm}
{
\small
\begin{lstlisting}[language=C,caption={\small{Code snippet of the \textsc{Gemm-Ncubed} kernel with its pragmas starting with ``\texttt{\#pragma}''. 
% Setting pragmas to different parameters leads to different designs with unique microarchitectures for which we aim to predict the quality in terms of latency and resource utilization on FPGA.
}},label={code:hls},floatplacement=H]
void gemm_N(double m1[4096],double m2[4096],double prod[4096])
{
  int i,j,k,k_col,i_col;
  double mult;
#pragma ACCEL PIPELINE auto{__PIPE__L0}
#pragma ACCEL TILE FACTOR=auto{__TILE__L0}
#pragma ACCEL PARALLEL FACTOR=auto{__PARA__L0}
for (i = 0; i < 64; i++) {    
#pragma ACCEL PIPELINE auto{__PIPE__L1}
#pragma ACCEL TILE FACTOR=auto{__TILE__L1}
#pragma ACCEL PARALLEL FACTOR=auto{__PARA__L1}
    for (j = 0; j < 64; j++) {
      i_col = i * 64;
      double sum = (double )0;
#pragma ACCEL PARALLEL reduction=sum FACTOR=auto{__PARA__L2}
      for (k = 0; k < 64; k++) {
        k_col = k * 64;
        mult = m1[i_col + k] * m2[k_col + j];
        sum += mult;
      }
      prod[i_col + j] = sum;
    }}}
\end{lstlisting}
\normalsize
}

\subsection{Training with Multiple Versions of Data}

\begin{figure}
    \centering
    \includegraphics[width=0.43\textwidth]{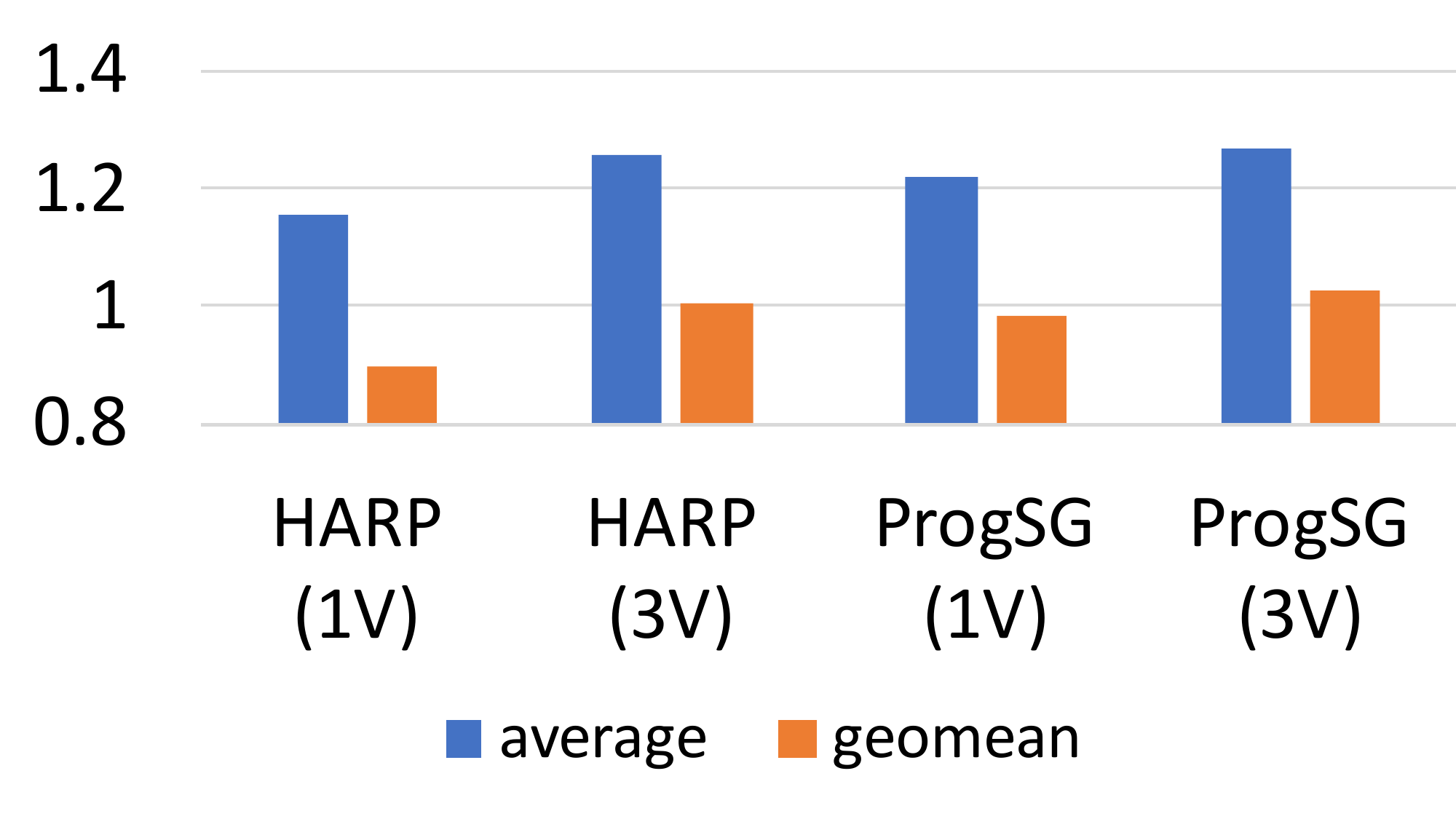}
    \caption{DSE results of HARP and ProgSG trained with 1 version (v21, denoted as 1V) and three versions (v18, v20, and v21, designated as 3V) of HLS tools.}
    \label{fig:fine-tune}
\end{figure}

In addition, \cdfg~\cite{sohrabizadeh2023robust} revealed that training the model with data obtained through multiple versions of HLS tools can improve the performance of the model. In their experiments, \cdfg is first trained with data of one version, then fine-tuned with data of another version. To investigate if the conclusion is true for \model, we conduct a similar experiment with data of three different versions using \cdfg and \model. Each model is first trained with data of the first version (HLS v18) for 1,000 epochs, then fine-tuned with the data of the second version (HLS v20) for 200 epochs, and finally fine-tuned with the data of third versions (HLS v21) for 400 epochs. Figure \ref{fig:fine-tune} illustrates the DSE performance of \cdfg and \model trained with 1 version and 3 versions of data. It is clear that training \model with multiple versions of data significantly increases its performance. 

\begin{figure}[htbp]
    \centering
    \begin{subfigure}[b]{0.23\textwidth}
        \centering
        \includegraphics[width=\textwidth,height=0.12\textheight]{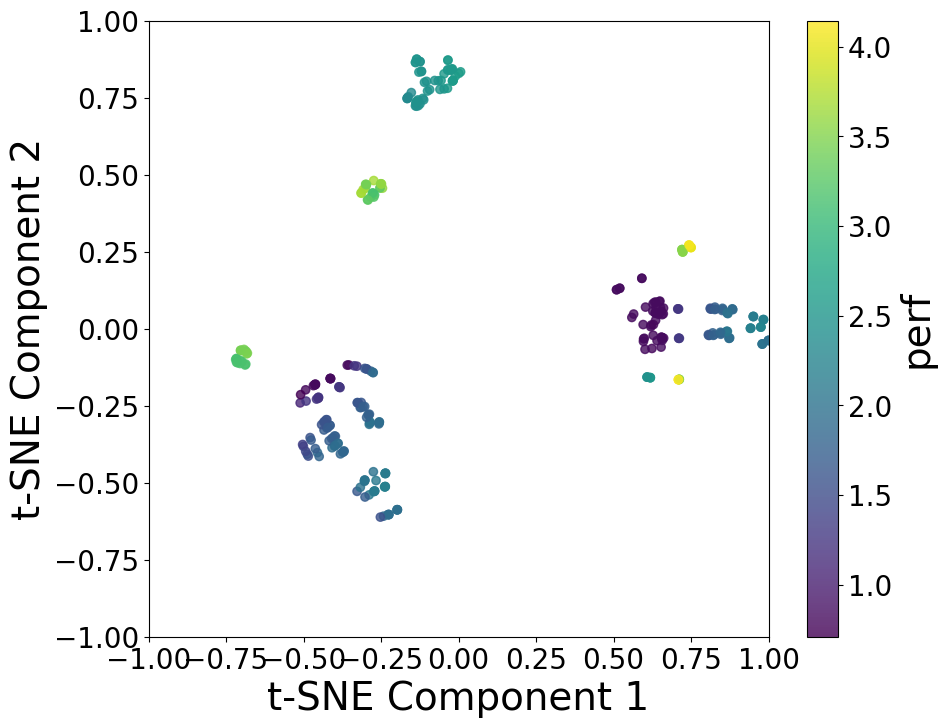}
        \caption{HARP embeddings}
        \label{fig:harp_vis}
    \end{subfigure}
    \hfill
    \begin{subfigure}[b]{0.23\textwidth}
        \centering
        \includegraphics[width=\textwidth,height=0.12\textheight]{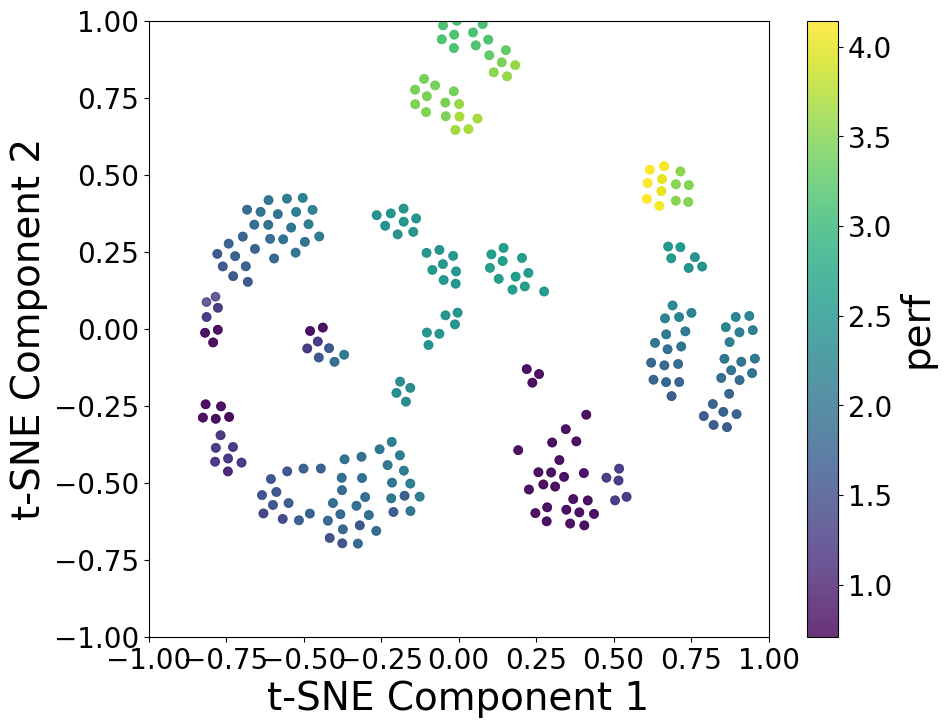}
        \caption{CodeT5 embeddings}
        \label{fig:codet5-vis}
    \end{subfigure}
    \hfill
    \begin{subfigure}[b]{0.23\textwidth}
        \centering
        \includegraphics[width=\textwidth,height=0.12\textheight]{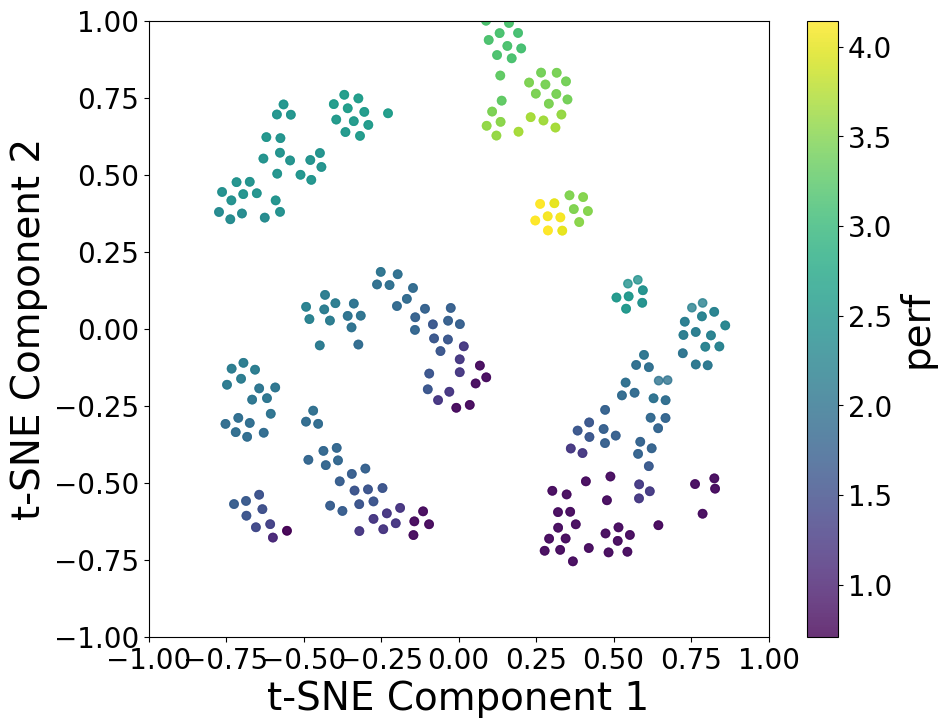}
        \caption{ProgSG embeddings}
        \label{fig:progsg-vis}
    \end{subfigure}
    \caption{Embedding visualizations with different methods for ``\textit{Correlation}'' kernel using t-SNE. The color indicates the value of ``perf'' target.}
    \label{fig:emb_vis_corr}
\end{figure}

\subsection{Embedding Visualization}

To gain further insight into why \model outperforms \codetfive and \cdfg, we visualize the embeddings of valid ``\textit{correlation}'' kernel designs in Figure \ref{fig:emb_vis_corr}. The colors represent the ground-truth performance targets. All methods form distinctive clusters with similar performance within each cluster. However, HARP's clusters are more crowded, likely due to the larger sizes of CodeT5 and \model, which can better differentiate designs. Additionally, \model's clusters align more closely with performance targets, as evidenced by the closer proximity of the purple points in \model's embeddings compared to those in \codetfive's embeddings. This suggests that \model's embeddings more accurately reflect design performance, thereby explaining its superior DSE and prediction results.

\section{Related Work} 
\label{sec-related}

\textbf{Machine Learning for Electronic Design Automation} 
\enspace
Machine learning (ML) for electronic design automation (EDA) is a rising research area~\citep{huang2021machine} with applications at various stages of hardware design, such as design verification~\citep{vasudevan2021learning,xu2020reinforcement,liang2023late}, high-level synthesis (HLS)~\citep{ustun2020accurate,sohrabizadeh2022automated,bai2022improving,wu2022ironman,sohrabizadeh2023robust,fu2023gpt4aigchip}, circuit design~\citep{ren2020paragraph,wang2022unsupervised,wang2020gcn,yang2022pre}, etc. This work focuses on obtaining representations of HLS designs using information from both the source code and the CDFG graph for FPGA design quality regression. Many works depict the input design/circuit as graphs~\citep{ustun2020accurate,ren2020paragraph,sohrabizadeh2022automated}. Recently, large language models (LLMs) are used to directly generate EDA scripts~\citep{liu2023chipnemo,liu2023verilogeval}. However, their results show that LLMs can only generate a few lines of scripts without considering the quality of the design. This work is among the first to combine both the source code and the graph modalities.

\textbf{Representation Learning for Programs}
\enspace
Based on the modality of data, current methods can be divided into source-code-based methods and data-structure-based methods. Source-code-based methods~\citep{cubert,feng2020codebert,wang2021codet5,svyatkovskiy2020intellicode} employ language models~\citep{devlin2018bert,radford2019language,raffel2020exploring,zheng2023codegeex,li2023starcoder,gunasekar2023textbooks,roziere2023code,fuhardware} on source code to perform various types of tasks.
However, it has not been demonstrated that these language models can predict the program's runtime, let alone predicting the corresponding hardware design performance.
The data-structure-based methods~\citep{alon2019code2vec,sohrabizadeh2022automated,sohrabizadeh2023robust} obtain the program embeddings from the data structure that represents a program. 
But the sizes of the models are usually small, restricting their prediction ability.




\textbf{Multi-modal Learning with Transformers}
\enspace
Modality-wise, transformers have been employed in cross-modality tasks spanning across vision~\citep{li2021align,wu2021hanet,huang2021learning}, language~\citep{zhang2019ernie,lee2022pix2struct}, source code~\citep{dai2022one}, knowledge graphs~\citep{yasunaga2022deep,rao2023retrieval}, audio~\citep{arandjelovic2017look,gan2020music}, point clouds~\citep{afham2022crosspoint}, etc. In fact, multi-modal learning using transformers has recently been considered possible for achieving generalist artificial intelligence~\citep{moor2023foundation,mai2023opportunities}.  
More thorough surveys on graphs and transformers can be found at~\citet{liu2023towards,li2023survey,jin2023large}. 

\greaseLM~\citep{zhang2022greaselm} combines GNN and transformer for knowledge-graph augmented QA task. It is similar to our task in that it also aims to combine the graph and the text modality. However, there are some differences in the targeted tasks that lead to distinct challenges and model design choices:
(1) The differences in program structures are subtle. Two designs with the same functionality can have a huge performance gap due to slight differences in the program or pragmas. So the global-level interaction in GreaseLM is not effective enough. A more fine-grained interaction between the modalities is needed.
(2) Efficiency is important to our task. As a larger inference overhead means slower design space exploration. Our program-derived graph is also much larger, with thousands of nodes and tokens. So the model needs to be efficient, forbidding the usage of full cross-attention between nodes and tokens.
(3) The programs are inherently hierarchical, making it possible to do interactions at multiple levels.
Given these unique challenges, we propose a novel model that interacts modalities at both global and block levels to maintain effectiveness while keeping it efficient. It is fair to say that we are the first to explore the cross-modality model for program representation learning.

\textbf{Graph Neural Networks Pre-training}
\enspace
\nop{
Most of the GNNs    \citep{hamilton2017inductive,kipf2016semi,DBLP:conf/iclr/VelickovicCCRLB18} fit into the message-passing framework where node representations are iteratively updated by aggregating the features of their neighbor nodes with a differentiable aggregation function. 
}
Existing self-supervised learning methods~\citep{xie2023self} can be divided into two categories: contrastive methods~\citep{sun2019infograph,You2020GraphCL,velickovic2018deep,qiu2020gcc} and predictive methods~\citep{xie2023self,kipf2016variational,gpt_gnn,s2grl,rong2020self,sun2020multi,hu2021rectifying}. 
To our knowledge, we are the first to explore pre-training GNNs with CDFGs. 

\section{Conclusion}
\label{sec-conc}


We propose \model, a novel two-modality program representation learning method for IC design (defined with HLS C/C++) optimization. The key assumption is that there is critical information in both the source code modality and the assembly code modality, which must be captured jointly. To achieve that, we propose a graph-summary-augmented sequence representation for the source code transformer, a fine-grained alignment utilization method, and a novel pre-training method for the GNN encoder for the CDFG. Experiments confirm the superiority of the proposed \model over baselines. We believe the core idea of using both modalities together with their alignment is general and can be adapted for other tasks.

\section{Acknowledgement}
This work was partially supported by NSF grants 2211557, 1937599,  2119643, and 2303037, SRC JUMP 2.0 PRISM Center, NASA, Okawa Foundation, Amazon Research, Cisco, Picsart, Snapchat, and the CDSC industrial partners (https://cdsc.ucla.edu/partners/). The authors would also like to thank Maria Brbic (EPFL) for early discussions on integrating GNN and LLM models, AMD/Xilinx for HACC equipment donation, and Marci Baun for editing the paper. J. Cong has a financial interest in AMD.

\bibliographystyle{ACM-Reference-Format}
\bibliography{bibliography}


\begin{thebibliography}{70}


\ifx \showCODEN    \undefined \def \showCODEN     #1{\unskip}     \fi
\ifx \showDOI      \undefined \def \showDOI       #1{#1}\fi
\ifx \showISBNx    \undefined \def \showISBNx     #1{\unskip}     \fi
\ifx \showISBNxiii \undefined \def \showISBNxiii  #1{\unskip}     \fi
\ifx \showISSN     \undefined \def \showISSN      #1{\unskip}     \fi
\ifx \showLCCN     \undefined \def \showLCCN      #1{\unskip}     \fi
\ifx \shownote     \undefined \def \shownote      #1{#1}          \fi
\ifx \showarticletitle \undefined \def \showarticletitle #1{#1}   \fi
\ifx \showURL      \undefined \def \showURL       {\relax}        \fi
\providecommand\bibfield[2]{#2}
\providecommand\bibinfo[2]{#2}
\providecommand\natexlab[1]{#1}
\providecommand\showeprint[2][]{arXiv:#2}

\bibitem[\protect\citeauthoryear{Afham, Dissanayake, Dissanayake, Dharmasiri, Thilakarathna, and Rodrigo}{Afham et~al\mbox{.}}{2022}]%
        {afham2022crosspoint}
\bibfield{author}{\bibinfo{person}{Mohamed Afham}, \bibinfo{person}{Isuru Dissanayake}, \bibinfo{person}{Dinithi Dissanayake}, \bibinfo{person}{Amaya Dharmasiri}, \bibinfo{person}{Kanchana Thilakarathna}, {and} \bibinfo{person}{Ranga Rodrigo}.} \bibinfo{year}{2022}\natexlab{}.
\newblock \showarticletitle{Crosspoint: Self-supervised cross-modal contrastive learning for 3d point cloud understanding}. In \bibinfo{booktitle}{\emph{Proceedings of the IEEE/CVF Conference on Computer Vision and Pattern Recognition}}. \bibinfo{pages}{9902--9912}.
\newblock


\bibitem[\protect\citeauthoryear{Alon, Zilberstein, Levy, and Yahav}{Alon et~al\mbox{.}}{2019}]%
        {alon2019code2vec}
\bibfield{author}{\bibinfo{person}{Uri Alon}, \bibinfo{person}{Meital Zilberstein}, \bibinfo{person}{Omer Levy}, {and} \bibinfo{person}{Eran Yahav}.} \bibinfo{year}{2019}\natexlab{}.
\newblock \showarticletitle{code2vec: Learning distributed representations of code}.
\newblock \bibinfo{journal}{\emph{Proceedings of the ACM on Programming Languages}} \bibinfo{volume}{3}, \bibinfo{number}{POPL} (\bibinfo{year}{2019}), \bibinfo{pages}{1--29}.
\newblock


\bibitem[\protect\citeauthoryear{Arandjelovic and Zisserman}{Arandjelovic and Zisserman}{2017}]%
        {arandjelovic2017look}
\bibfield{author}{\bibinfo{person}{Relja Arandjelovic} {and} \bibinfo{person}{Andrew Zisserman}.} \bibinfo{year}{2017}\natexlab{}.
\newblock \showarticletitle{Look, listen and learn}. In \bibinfo{booktitle}{\emph{Proceedings of the IEEE international conference on computer vision}}. \bibinfo{pages}{609--617}.
\newblock


\bibitem[\protect\citeauthoryear{Bai, Sohrabizadeh, Qin, Hu, Sun, and Cong}{Bai et~al\mbox{.}}{2023}]%
        {bai2023towards}
\bibfield{author}{\bibinfo{person}{Yunsheng Bai}, \bibinfo{person}{Atefeh Sohrabizadeh}, \bibinfo{person}{Zongyue Qin}, \bibinfo{person}{Ziniu Hu}, \bibinfo{person}{Yizhou Sun}, {and} \bibinfo{person}{Jason Cong}.} \bibinfo{year}{2023}\natexlab{}.
\newblock \showarticletitle{Towards a Comprehensive Benchmark for High-Level Synthesis Targeted to FPGAs}. In \bibinfo{booktitle}{\emph{Thirty-seventh Conference on Neural Information Processing Systems Datasets and Benchmarks Track}}.
\newblock


\bibitem[\protect\citeauthoryear{Bai, Sohrabizadeh, Sun, and Cong}{Bai et~al\mbox{.}}{2022}]%
        {bai2022improving}
\bibfield{author}{\bibinfo{person}{Yunsheng Bai}, \bibinfo{person}{Atefeh Sohrabizadeh}, \bibinfo{person}{Yizhou Sun}, {and} \bibinfo{person}{Jason Cong}.} \bibinfo{year}{2022}\natexlab{}.
\newblock \showarticletitle{Improving GNN-based accelerator design automation with meta learning}. In \bibinfo{booktitle}{\emph{Proceedings of the 59th ACM/IEEE Design Automation Conference}}. \bibinfo{pages}{1347--1350}.
\newblock


\bibitem[\protect\citeauthoryear{Clevert, Unterthiner, and Hochreiter}{Clevert et~al\mbox{.}}{2015}]%
        {clevert2015fast}
\bibfield{author}{\bibinfo{person}{Djork-Arn{\'e} Clevert}, \bibinfo{person}{Thomas Unterthiner}, {and} \bibinfo{person}{Sepp Hochreiter}.} \bibinfo{year}{2015}\natexlab{}.
\newblock \showarticletitle{Fast and accurate deep network learning by exponential linear units (elus)}.
\newblock \bibinfo{journal}{\emph{arXiv preprint arXiv:1511.07289}} (\bibinfo{year}{2015}).
\newblock


\bibitem[\protect\citeauthoryear{Cummins, Fisches, Ben-Nun, Hoefler, O’Boyle, and Leather}{Cummins et~al\mbox{.}}{2021}]%
        {cummins2021programl}
\bibfield{author}{\bibinfo{person}{Chris Cummins}, \bibinfo{person}{Zacharias~V Fisches}, \bibinfo{person}{Tal Ben-Nun}, \bibinfo{person}{Torsten Hoefler}, \bibinfo{person}{Michael~FP O’Boyle}, {and} \bibinfo{person}{Hugh Leather}.} \bibinfo{year}{2021}\natexlab{}.
\newblock \showarticletitle{Programl: A graph-based program representation for data flow analysis and compiler optimizations}. In \bibinfo{booktitle}{\emph{International Conference on Machine Learning}}. PMLR, \bibinfo{pages}{2244--2253}.
\newblock


\bibitem[\protect\citeauthoryear{Dai, Tang, Liu, Tan, Zhou, Wang, Feng, Zhang, Hu, and Shi}{Dai et~al\mbox{.}}{2022}]%
        {dai2022one}
\bibfield{author}{\bibinfo{person}{Yong Dai}, \bibinfo{person}{Duyu Tang}, \bibinfo{person}{Liangxin Liu}, \bibinfo{person}{Minghuan Tan}, \bibinfo{person}{Cong Zhou}, \bibinfo{person}{Jingquan Wang}, \bibinfo{person}{Zhangyin Feng}, \bibinfo{person}{Fan Zhang}, \bibinfo{person}{Xueyu Hu}, {and} \bibinfo{person}{Shuming Shi}.} \bibinfo{year}{2022}\natexlab{}.
\newblock \showarticletitle{One model, multiple modalities: A sparsely activated approach for text, sound, image, video and code}.
\newblock \bibinfo{journal}{\emph{arXiv preprint arXiv:2205.06126}} (\bibinfo{year}{2022}).
\newblock


\bibitem[\protect\citeauthoryear{Devlin, Chang, Lee, and Toutanova}{Devlin et~al\mbox{.}}{2018}]%
        {devlin2018bert}
\bibfield{author}{\bibinfo{person}{Jacob Devlin}, \bibinfo{person}{Ming-Wei Chang}, \bibinfo{person}{Kenton Lee}, {and} \bibinfo{person}{Kristina Toutanova}.} \bibinfo{year}{2018}\natexlab{}.
\newblock \showarticletitle{Bert: Pre-training of deep bidirectional transformers for language understanding}.
\newblock \bibinfo{journal}{\emph{arXiv preprint arXiv:1810.04805}} (\bibinfo{year}{2018}).
\newblock


\bibitem[\protect\citeauthoryear{Feng, Guo, Tang, Duan, Feng, Gong, Shou, Qin, Liu, Jiang, et~al\mbox{.}}{Feng et~al\mbox{.}}{2020}]%
        {feng2020codebert}
\bibfield{author}{\bibinfo{person}{Zhangyin Feng}, \bibinfo{person}{Daya Guo}, \bibinfo{person}{Duyu Tang}, \bibinfo{person}{Nan Duan}, \bibinfo{person}{Xiaocheng Feng}, \bibinfo{person}{Ming Gong}, \bibinfo{person}{Linjun Shou}, \bibinfo{person}{Bing Qin}, \bibinfo{person}{Ting Liu}, \bibinfo{person}{Daxin Jiang}, {et~al\mbox{.}}} \bibinfo{year}{2020}\natexlab{}.
\newblock \showarticletitle{Codebert: A pre-trained model for programming and natural languages}.
\newblock \bibinfo{journal}{\emph{arXiv preprint arXiv:2002.08155}} (\bibinfo{year}{2020}).
\newblock


\bibitem[\protect\citeauthoryear{Fu, Li, Zhao, Ma, Dutta, Zhang, Yang, Jin, and Guo}{Fu et~al\mbox{.}}{2024}]%
        {fuhardware}
\bibfield{author}{\bibinfo{person}{Weimin Fu}, \bibinfo{person}{Shijie Li}, \bibinfo{person}{Yifang Zhao}, \bibinfo{person}{Haocheng Ma}, \bibinfo{person}{Raj Dutta}, \bibinfo{person}{Xuan Zhang}, \bibinfo{person}{Kaichen Yang}, \bibinfo{person}{Yier Jin}, {and} \bibinfo{person}{Xiaolong Guo}.} \bibinfo{year}{2024}\natexlab{}.
\newblock \showarticletitle{Hardware Phi-1.5 B: A Large Language Model Encodes Hardware Domain Specific Knowledge}.
\newblock \bibinfo{journal}{\emph{arXiv preprint arXiv:2402.01728}} (\bibinfo{year}{2024}).
\newblock


\bibitem[\protect\citeauthoryear{Fu, Zhang, Yu, Li, Ye, Li, Wan, and Lin}{Fu et~al\mbox{.}}{2023}]%
        {fu2023gpt4aigchip}
\bibfield{author}{\bibinfo{person}{Yonggan Fu}, \bibinfo{person}{Yongan Zhang}, \bibinfo{person}{Zhongzhi Yu}, \bibinfo{person}{Sixu Li}, \bibinfo{person}{Zhifan Ye}, \bibinfo{person}{Chaojian Li}, \bibinfo{person}{Cheng Wan}, {and} \bibinfo{person}{Yingyan~Celine Lin}.} \bibinfo{year}{2023}\natexlab{}.
\newblock \showarticletitle{Gpt4aigchip: Towards next-generation ai accelerator design automation via large language models}. In \bibinfo{booktitle}{\emph{2023 IEEE/ACM International Conference on Computer Aided Design (ICCAD)}}. IEEE, \bibinfo{pages}{1--9}.
\newblock


\bibitem[\protect\citeauthoryear{Gan, Huang, Zhao, Tenenbaum, and Torralba}{Gan et~al\mbox{.}}{2020}]%
        {gan2020music}
\bibfield{author}{\bibinfo{person}{Chuang Gan}, \bibinfo{person}{Deng Huang}, \bibinfo{person}{Hang Zhao}, \bibinfo{person}{Joshua~B Tenenbaum}, {and} \bibinfo{person}{Antonio Torralba}.} \bibinfo{year}{2020}\natexlab{}.
\newblock \showarticletitle{Music gesture for visual sound separation}. In \bibinfo{booktitle}{\emph{Proceedings of the IEEE/CVF Conference on Computer Vision and Pattern Recognition}}. \bibinfo{pages}{10478--10487}.
\newblock


\bibitem[\protect\citeauthoryear{Gunasekar, Zhang, Aneja, Mendes, Del~Giorno, Gopi, Javaheripi, Kauffmann, de~Rosa, Saarikivi, et~al\mbox{.}}{Gunasekar et~al\mbox{.}}{2023}]%
        {gunasekar2023textbooks}
\bibfield{author}{\bibinfo{person}{Suriya Gunasekar}, \bibinfo{person}{Yi Zhang}, \bibinfo{person}{Jyoti Aneja}, \bibinfo{person}{Caio C{\'e}sar~Teodoro Mendes}, \bibinfo{person}{Allie Del~Giorno}, \bibinfo{person}{Sivakanth Gopi}, \bibinfo{person}{Mojan Javaheripi}, \bibinfo{person}{Piero Kauffmann}, \bibinfo{person}{Gustavo de Rosa}, \bibinfo{person}{Olli Saarikivi}, {et~al\mbox{.}}} \bibinfo{year}{2023}\natexlab{}.
\newblock \showarticletitle{Textbooks Are All You Need}.
\newblock \bibinfo{journal}{\emph{arXiv preprint arXiv:2306.11644}} (\bibinfo{year}{2023}).
\newblock


\bibitem[\protect\citeauthoryear{Guo, Ren, Lu, Feng, Tang, Liu, Zhou, Duan, Svyatkovskiy, Fu, et~al\mbox{.}}{Guo et~al\mbox{.}}{2020}]%
        {guo2020graphcodebert}
\bibfield{author}{\bibinfo{person}{Daya Guo}, \bibinfo{person}{Shuo Ren}, \bibinfo{person}{Shuai Lu}, \bibinfo{person}{Zhangyin Feng}, \bibinfo{person}{Duyu Tang}, \bibinfo{person}{Shujie Liu}, \bibinfo{person}{Long Zhou}, \bibinfo{person}{Nan Duan}, \bibinfo{person}{Alexey Svyatkovskiy}, \bibinfo{person}{Shengyu Fu}, {et~al\mbox{.}}} \bibinfo{year}{2020}\natexlab{}.
\newblock \showarticletitle{Graphcodebert: Pre-training code representations with data flow}.
\newblock \bibinfo{journal}{\emph{arXiv preprint arXiv:2009.08366}} (\bibinfo{year}{2020}).
\newblock


\bibitem[\protect\citeauthoryear{Hu, Dong, Wang, Chang, and Sun}{Hu et~al\mbox{.}}{2020}]%
        {gpt_gnn}
\bibfield{author}{\bibinfo{person}{Ziniu Hu}, \bibinfo{person}{Yuxiao Dong}, \bibinfo{person}{Kuansan Wang}, \bibinfo{person}{Kai-Wei Chang}, {and} \bibinfo{person}{Yizhou Sun}.} \bibinfo{year}{2020}\natexlab{}.
\newblock \showarticletitle{GPT-GNN: Generative Pre-Training of Graph Neural Networks}. In \bibinfo{booktitle}{\emph{Proceedings of the 26th ACM SIGKDD Conference on Knowledge Discovery and Data Mining}}.
\newblock


\bibitem[\protect\citeauthoryear{Hu, Kou, Zhang, Li, Yang, and Liu}{Hu et~al\mbox{.}}{2021}]%
        {hu2021rectifying}
\bibfield{author}{\bibinfo{person}{Zhihui Hu}, \bibinfo{person}{Guang Kou}, \bibinfo{person}{Haoyu Zhang}, \bibinfo{person}{Na Li}, \bibinfo{person}{Ke Yang}, {and} \bibinfo{person}{Lin Liu}.} \bibinfo{year}{2021}\natexlab{}.
\newblock \showarticletitle{Rectifying Pseudo Labels: Iterative Feature Clustering for Graph Representation Learning}. In \bibinfo{booktitle}{\emph{Proceedings of the 30th ACM International Conference on Information and Knowledge Management}} (Virtual Event, Queensland, Australia) \emph{(\bibinfo{series}{CIKM '21})}. \bibinfo{publisher}{Association for Computing Machinery}, \bibinfo{address}{New York, NY, USA}, \bibinfo{pages}{720–729}.
\newblock
\showISBNx{9781450384469}
\urldef\tempurl%
\url{https://doi.org/10.1145/3459637.3482469}
\showDOI{\tempurl}


\bibitem[\protect\citeauthoryear{Huang, Hu, He, Liu, Ma, Shen, Wu, Xu, Zhang, Zhong, et~al\mbox{.}}{Huang et~al\mbox{.}}{2021a}]%
        {huang2021machine}
\bibfield{author}{\bibinfo{person}{Guyue Huang}, \bibinfo{person}{Jingbo Hu}, \bibinfo{person}{Yifan He}, \bibinfo{person}{Jialong Liu}, \bibinfo{person}{Mingyuan Ma}, \bibinfo{person}{Zhaoyang Shen}, \bibinfo{person}{Juejian Wu}, \bibinfo{person}{Yuanfan Xu}, \bibinfo{person}{Hengrui Zhang}, \bibinfo{person}{Kai Zhong}, {et~al\mbox{.}}} \bibinfo{year}{2021}\natexlab{a}.
\newblock \showarticletitle{Machine learning for electronic design automation: A survey}.
\newblock \bibinfo{journal}{\emph{ACM Transactions on Design Automation of Electronic Systems (TODAES)}} \bibinfo{volume}{26}, \bibinfo{number}{5} (\bibinfo{year}{2021}), \bibinfo{pages}{1--46}.
\newblock


\bibitem[\protect\citeauthoryear{Huang, Niu, Liu, Ding, Xiao, Wu, and Peng}{Huang et~al\mbox{.}}{2021b}]%
        {huang2021learning}
\bibfield{author}{\bibinfo{person}{Zhenyu Huang}, \bibinfo{person}{Guocheng Niu}, \bibinfo{person}{Xiao Liu}, \bibinfo{person}{Wenbiao Ding}, \bibinfo{person}{Xinyan Xiao}, \bibinfo{person}{Hua Wu}, {and} \bibinfo{person}{Xi Peng}.} \bibinfo{year}{2021}\natexlab{b}.
\newblock \showarticletitle{Learning with noisy correspondence for cross-modal matching}.
\newblock \bibinfo{journal}{\emph{NeurIPS}}  \bibinfo{volume}{34} (\bibinfo{year}{2021}), \bibinfo{pages}{29406--29419}.
\newblock


\bibitem[\protect\citeauthoryear{Jin, Liu, Han, Jiang, Ji, and Han}{Jin et~al\mbox{.}}{2023}]%
        {jin2023large}
\bibfield{author}{\bibinfo{person}{Bowen Jin}, \bibinfo{person}{Gang Liu}, \bibinfo{person}{Chi Han}, \bibinfo{person}{Meng Jiang}, \bibinfo{person}{Heng Ji}, {and} \bibinfo{person}{Jiawei Han}.} \bibinfo{year}{2023}\natexlab{}.
\newblock \showarticletitle{Large Language Models on Graphs: A Comprehensive Survey}.
\newblock \bibinfo{journal}{\emph{arXiv preprint arXiv:2312.02783}} (\bibinfo{year}{2023}).
\newblock


\bibitem[\protect\citeauthoryear{Kanade, Maniatis, Balakrishnan, and Shi}{Kanade et~al\mbox{.}}{2020}]%
        {cubert}
\bibfield{author}{\bibinfo{person}{Aditya Kanade}, \bibinfo{person}{Petros Maniatis}, \bibinfo{person}{Gogul Balakrishnan}, {and} \bibinfo{person}{Kensen Shi}.} \bibinfo{year}{2020}\natexlab{}.
\newblock \showarticletitle{Learning and Evaluating Contextual Embedding of Source Code}. In \bibinfo{booktitle}{\emph{Proceedings of the 37th International Conference on Machine Learning}} \emph{(\bibinfo{series}{Proceedings of Machine Learning Research})}, \bibfield{editor}{\bibinfo{person}{Hal~Daumé III} {and} \bibinfo{person}{Aarti Singh}} (Eds.), Vol.~\bibinfo{volume}{119}. \bibinfo{publisher}{PMLR}, \bibinfo{pages}{5110--5121}.
\newblock
\urldef\tempurl%
\url{https://proceedings.mlr.press/v119/kanade20a.html}
\showURL{%
\tempurl}


\bibitem[\protect\citeauthoryear{Kipf and Welling}{Kipf and Welling}{2016}]%
        {kipf2016variational}
\bibfield{author}{\bibinfo{person}{Thomas~N Kipf} {and} \bibinfo{person}{Max Welling}.} \bibinfo{year}{2016}\natexlab{}.
\newblock \showarticletitle{Variational graph auto-encoders}.
\newblock \bibinfo{journal}{\emph{NIPS Workshop on Bayesian Deep Learning}} (\bibinfo{year}{2016}).
\newblock


\bibitem[\protect\citeauthoryear{Lattner and Adve}{Lattner and Adve}{2004}]%
        {lattner2004llvm}
\bibfield{author}{\bibinfo{person}{Chris Lattner} {and} \bibinfo{person}{Vikram Adve}.} \bibinfo{year}{2004}\natexlab{}.
\newblock \showarticletitle{LLVM: A compilation framework for lifelong program analysis \& transformation}. In \bibinfo{booktitle}{\emph{International Symposium on CGO}}.
\newblock


\bibitem[\protect\citeauthoryear{Lee, Joshi, Turc, Hu, Liu, Eisenschlos, Khandelwal, Shaw, Chang, and Toutanova}{Lee et~al\mbox{.}}{2022}]%
        {lee2022pix2struct}
\bibfield{author}{\bibinfo{person}{Kenton Lee}, \bibinfo{person}{Mandar Joshi}, \bibinfo{person}{Iulia Turc}, \bibinfo{person}{Hexiang Hu}, \bibinfo{person}{Fangyu Liu}, \bibinfo{person}{Julian Eisenschlos}, \bibinfo{person}{Urvashi Khandelwal}, \bibinfo{person}{Peter Shaw}, \bibinfo{person}{Ming-Wei Chang}, {and} \bibinfo{person}{Kristina Toutanova}.} \bibinfo{year}{2022}\natexlab{}.
\newblock \showarticletitle{Pix2Struct: Screenshot parsing as pretraining for visual language understanding}.
\newblock \bibinfo{journal}{\emph{arXiv preprint arXiv:2210.03347}} (\bibinfo{year}{2022}).
\newblock


\bibitem[\protect\citeauthoryear{Li, Selvaraju, Gotmare, Joty, Xiong, and Hoi}{Li et~al\mbox{.}}{2021}]%
        {li2021align}
\bibfield{author}{\bibinfo{person}{Junnan Li}, \bibinfo{person}{Ramprasaath Selvaraju}, \bibinfo{person}{Akhilesh Gotmare}, \bibinfo{person}{Shafiq Joty}, \bibinfo{person}{Caiming Xiong}, {and} \bibinfo{person}{Steven Chu~Hong Hoi}.} \bibinfo{year}{2021}\natexlab{}.
\newblock \showarticletitle{Align before fuse: Vision and language representation learning with momentum distillation}.
\newblock \bibinfo{journal}{\emph{NeurIPS}}  \bibinfo{volume}{34} (\bibinfo{year}{2021}), \bibinfo{pages}{9694--9705}.
\newblock


\bibitem[\protect\citeauthoryear{Li, Allal, Zi, Muennighoff, Kocetkov, Mou, Marone, Akiki, Li, Chim, et~al\mbox{.}}{Li et~al\mbox{.}}{2023a}]%
        {li2023starcoder}
\bibfield{author}{\bibinfo{person}{Raymond Li}, \bibinfo{person}{Loubna~Ben Allal}, \bibinfo{person}{Yangtian Zi}, \bibinfo{person}{Niklas Muennighoff}, \bibinfo{person}{Denis Kocetkov}, \bibinfo{person}{Chenghao Mou}, \bibinfo{person}{Marc Marone}, \bibinfo{person}{Christopher Akiki}, \bibinfo{person}{Jia Li}, \bibinfo{person}{Jenny Chim}, {et~al\mbox{.}}} \bibinfo{year}{2023}\natexlab{a}.
\newblock \showarticletitle{StarCoder: may the source be with you!}
\newblock \bibinfo{journal}{\emph{arXiv preprint arXiv:2305.06161}} (\bibinfo{year}{2023}).
\newblock


\bibitem[\protect\citeauthoryear{Li, Li, Wang, Li, Sun, Cheng, and Yu}{Li et~al\mbox{.}}{2023b}]%
        {li2023survey}
\bibfield{author}{\bibinfo{person}{Yuhan Li}, \bibinfo{person}{Zhixun Li}, \bibinfo{person}{Peisong Wang}, \bibinfo{person}{Jia Li}, \bibinfo{person}{Xiangguo Sun}, \bibinfo{person}{Hong Cheng}, {and} \bibinfo{person}{Jeffrey~Xu Yu}.} \bibinfo{year}{2023}\natexlab{b}.
\newblock \showarticletitle{A survey of graph meets large language model: Progress and future directions}.
\newblock \bibinfo{journal}{\emph{arXiv preprint arXiv:2311.12399}} (\bibinfo{year}{2023}).
\newblock


\bibitem[\protect\citeauthoryear{Liang, Pinckney, Chai, Ren, and Khailany}{Liang et~al\mbox{.}}{2023}]%
        {liang2023late}
\bibfield{author}{\bibinfo{person}{Rongjian Liang}, \bibinfo{person}{Nathaniel Pinckney}, \bibinfo{person}{Yuji Chai}, \bibinfo{person}{Haoxin Ren}, {and} \bibinfo{person}{Brucek Khailany}.} \bibinfo{year}{2023}\natexlab{}.
\newblock \showarticletitle{Late Breaking Results: Test Selection For RTL Coverage By Unsupervised Learning From Fast Functional Simulation}. In \bibinfo{booktitle}{\emph{2023 60th ACM/IEEE Design Automation Conference (DAC)}}. IEEE, \bibinfo{pages}{1--2}.
\newblock


\bibitem[\protect\citeauthoryear{Liu, Yang, Lu, Chen, Li, Zhang, Bai, Fang, Sun, Yu, et~al\mbox{.}}{Liu et~al\mbox{.}}{2023c}]%
        {liu2023towards}
\bibfield{author}{\bibinfo{person}{Jiawei Liu}, \bibinfo{person}{Cheng Yang}, \bibinfo{person}{Zhiyuan Lu}, \bibinfo{person}{Junze Chen}, \bibinfo{person}{Yibo Li}, \bibinfo{person}{Mengmei Zhang}, \bibinfo{person}{Ting Bai}, \bibinfo{person}{Yuan Fang}, \bibinfo{person}{Lichao Sun}, \bibinfo{person}{Philip~S Yu}, {et~al\mbox{.}}} \bibinfo{year}{2023}\natexlab{c}.
\newblock \showarticletitle{Towards graph foundation models: A survey and beyond}.
\newblock \bibinfo{journal}{\emph{arXiv preprint arXiv:2310.11829}} (\bibinfo{year}{2023}).
\newblock


\bibitem[\protect\citeauthoryear{Liu, Ene, Kirby, Cheng, Pinckney, Liang, Alben, Anand, Banerjee, Bayraktaroglu, et~al\mbox{.}}{Liu et~al\mbox{.}}{2023a}]%
        {liu2023chipnemo}
\bibfield{author}{\bibinfo{person}{Mingjie Liu}, \bibinfo{person}{Teodor-Dumitru Ene}, \bibinfo{person}{Robert Kirby}, \bibinfo{person}{Chris Cheng}, \bibinfo{person}{Nathaniel Pinckney}, \bibinfo{person}{Rongjian Liang}, \bibinfo{person}{Jonah Alben}, \bibinfo{person}{Himyanshu Anand}, \bibinfo{person}{Sanmitra Banerjee}, \bibinfo{person}{Ismet Bayraktaroglu}, {et~al\mbox{.}}} \bibinfo{year}{2023}\natexlab{a}.
\newblock \showarticletitle{Chipnemo: Domain-adapted llms for chip design}.
\newblock \bibinfo{journal}{\emph{arXiv preprint arXiv:2311.00176}} (\bibinfo{year}{2023}).
\newblock


\bibitem[\protect\citeauthoryear{Liu, Pinckney, Khailany, and Ren}{Liu et~al\mbox{.}}{2023b}]%
        {liu2023verilogeval}
\bibfield{author}{\bibinfo{person}{Mingjie Liu}, \bibinfo{person}{Nathaniel Pinckney}, \bibinfo{person}{Brucek Khailany}, {and} \bibinfo{person}{Haoxing Ren}.} \bibinfo{year}{2023}\natexlab{b}.
\newblock \showarticletitle{Verilogeval: Evaluating large language models for verilog code generation}. In \bibinfo{booktitle}{\emph{2023 IEEE/ACM International Conference on Computer Aided Design (ICCAD)}}. IEEE, \bibinfo{pages}{1--8}.
\newblock


\bibitem[\protect\citeauthoryear{Loshchilov and Hutter}{Loshchilov and Hutter}{2019}]%
        {loshchilov2017decoupled}
\bibfield{author}{\bibinfo{person}{Ilya Loshchilov} {and} \bibinfo{person}{Frank Hutter}.} \bibinfo{year}{2019}\natexlab{}.
\newblock \showarticletitle{Decoupled weight decay regularization}.
\newblock \bibinfo{journal}{\emph{ICLR}} (\bibinfo{year}{2019}).
\newblock


\bibitem[\protect\citeauthoryear{Mai, Huang, Sun, Song, Mishra, Liu, Gao, Liu, Cong, Hu, et~al\mbox{.}}{Mai et~al\mbox{.}}{2023}]%
        {mai2023opportunities}
\bibfield{author}{\bibinfo{person}{Gengchen Mai}, \bibinfo{person}{Weiming Huang}, \bibinfo{person}{Jin Sun}, \bibinfo{person}{Suhang Song}, \bibinfo{person}{Deepak Mishra}, \bibinfo{person}{Ninghao Liu}, \bibinfo{person}{Song Gao}, \bibinfo{person}{Tianming Liu}, \bibinfo{person}{Gao Cong}, \bibinfo{person}{Yingjie Hu}, {et~al\mbox{.}}} \bibinfo{year}{2023}\natexlab{}.
\newblock \showarticletitle{On the opportunities and challenges of foundation models for geospatial artificial intelligence}.
\newblock \bibinfo{journal}{\emph{arXiv preprint arXiv:2304.06798}} (\bibinfo{year}{2023}).
\newblock


\bibitem[\protect\citeauthoryear{Moor, Banerjee, Abad, Krumholz, Leskovec, Topol, and Rajpurkar}{Moor et~al\mbox{.}}{2023}]%
        {moor2023foundation}
\bibfield{author}{\bibinfo{person}{Michael Moor}, \bibinfo{person}{Oishi Banerjee}, \bibinfo{person}{Zahra Shakeri~Hossein Abad}, \bibinfo{person}{Harlan~M Krumholz}, \bibinfo{person}{Jure Leskovec}, \bibinfo{person}{Eric~J Topol}, {and} \bibinfo{person}{Pranav Rajpurkar}.} \bibinfo{year}{2023}\natexlab{}.
\newblock \showarticletitle{Foundation models for generalist medical artificial intelligence}.
\newblock \bibinfo{journal}{\emph{Nature}} \bibinfo{volume}{616}, \bibinfo{number}{7956} (\bibinfo{year}{2023}), \bibinfo{pages}{259--265}.
\newblock


\bibitem[\protect\citeauthoryear{Peng, Dong, Luo, Wu, and Zheng}{Peng et~al\mbox{.}}{2020}]%
        {s2grl}
\bibfield{author}{\bibinfo{person}{Zhen Peng}, \bibinfo{person}{Yixiang Dong}, \bibinfo{person}{Minnan Luo}, \bibinfo{person}{Xiao-Ming Wu}, {and} \bibinfo{person}{Qinghua Zheng}.} \bibinfo{year}{2020}\natexlab{}.
\newblock \showarticletitle{Self-Supervised Graph Representation Learning via Global Context Prediction}.
\newblock


\bibitem[\protect\citeauthoryear{Qiu, Chen, Dong, Zhang, Yang, Ding, Wang, and Tang}{Qiu et~al\mbox{.}}{2020}]%
        {qiu2020gcc}
\bibfield{author}{\bibinfo{person}{Jiezhong Qiu}, \bibinfo{person}{Qibin Chen}, \bibinfo{person}{Yuxiao Dong}, \bibinfo{person}{Jing Zhang}, \bibinfo{person}{Hongxia Yang}, \bibinfo{person}{Ming Ding}, \bibinfo{person}{Kuansan Wang}, {and} \bibinfo{person}{Jie Tang}.} \bibinfo{year}{2020}\natexlab{}.
\newblock \showarticletitle{GCC: Graph Contrastive Coding for Graph Neural Network Pre-Training}.
\newblock \bibinfo{journal}{\emph{arXiv preprint arXiv:2006.09963}} (\bibinfo{year}{2020}).
\newblock


\bibitem[\protect\citeauthoryear{Radford, Wu, Child, Luan, Amodei, Sutskever, et~al\mbox{.}}{Radford et~al\mbox{.}}{2019}]%
        {radford2019language}
\bibfield{author}{\bibinfo{person}{Alec Radford}, \bibinfo{person}{Jeffrey Wu}, \bibinfo{person}{Rewon Child}, \bibinfo{person}{David Luan}, \bibinfo{person}{Dario Amodei}, \bibinfo{person}{Ilya Sutskever}, {et~al\mbox{.}}} \bibinfo{year}{2019}\natexlab{}.
\newblock \showarticletitle{Language models are unsupervised multitask learners}.
\newblock \bibinfo{journal}{\emph{OpenAI blog}} \bibinfo{volume}{1}, \bibinfo{number}{8} (\bibinfo{year}{2019}), \bibinfo{pages}{9}.
\newblock


\bibitem[\protect\citeauthoryear{Raffel, Shazeer, Roberts, Lee, Narang, Matena, Zhou, Li, and Liu}{Raffel et~al\mbox{.}}{2020}]%
        {raffel2020exploring}
\bibfield{author}{\bibinfo{person}{Colin Raffel}, \bibinfo{person}{Noam Shazeer}, \bibinfo{person}{Adam Roberts}, \bibinfo{person}{Katherine Lee}, \bibinfo{person}{Sharan Narang}, \bibinfo{person}{Michael Matena}, \bibinfo{person}{Yanqi Zhou}, \bibinfo{person}{Wei Li}, {and} \bibinfo{person}{Peter~J Liu}.} \bibinfo{year}{2020}\natexlab{}.
\newblock \showarticletitle{Exploring the limits of transfer learning with a unified text-to-text transformer}.
\newblock \bibinfo{journal}{\emph{The Journal of Machine Learning Research}} \bibinfo{volume}{21}, \bibinfo{number}{1} (\bibinfo{year}{2020}), \bibinfo{pages}{5485--5551}.
\newblock


\bibitem[\protect\citeauthoryear{Rao, Shan, Liu, Zhou, and Yang}{Rao et~al\mbox{.}}{2023}]%
        {rao2023retrieval}
\bibfield{author}{\bibinfo{person}{Jiahua Rao}, \bibinfo{person}{Zifei Shan}, \bibinfo{person}{Longpo Liu}, \bibinfo{person}{Yao Zhou}, {and} \bibinfo{person}{Yuedong Yang}.} \bibinfo{year}{2023}\natexlab{}.
\newblock \showarticletitle{Retrieval-based Knowledge Augmented Vision Language Pre-training}.
\newblock \bibinfo{journal}{\emph{arXiv preprint arXiv:2304.13923}} (\bibinfo{year}{2023}).
\newblock


\bibitem[\protect\citeauthoryear{Reagen, Adolf, Shao, Wei, and Brooks}{Reagen et~al\mbox{.}}{2014}]%
        {machsuite}
\bibfield{author}{\bibinfo{person}{Brandon Reagen}, \bibinfo{person}{Robert Adolf}, \bibinfo{person}{Yakun~Sophia Shao}, \bibinfo{person}{Gu-Yeon Wei}, {and} \bibinfo{person}{David Brooks}.} \bibinfo{year}{2014}\natexlab{}.
\newblock \showarticletitle{Machsuite: Benchmarks for accelerator design and customized architectures}. In \bibinfo{booktitle}{\emph{IISWC}}.
\newblock


\bibitem[\protect\citeauthoryear{Ren, Kokai, Turner, and Ku}{Ren et~al\mbox{.}}{2020}]%
        {ren2020paragraph}
\bibfield{author}{\bibinfo{person}{Haoxing Ren}, \bibinfo{person}{George~F Kokai}, \bibinfo{person}{Walker~J Turner}, {and} \bibinfo{person}{Ting-Sheng Ku}.} \bibinfo{year}{2020}\natexlab{}.
\newblock \showarticletitle{ParaGraph: Layout parasitics and device parameter prediction using graph neural networks}. In \bibinfo{booktitle}{\emph{2020 57th ACM/IEEE Design Automation Conference (DAC)}}. IEEE, \bibinfo{pages}{1--6}.
\newblock


\bibitem[\protect\citeauthoryear{Rong, Bian, Xu, Xie, Wei, Huang, and Huang}{Rong et~al\mbox{.}}{2020}]%
        {rong2020self}
\bibfield{author}{\bibinfo{person}{Yu Rong}, \bibinfo{person}{Yatao Bian}, \bibinfo{person}{Tingyang Xu}, \bibinfo{person}{Weiyang Xie}, \bibinfo{person}{Ying Wei}, \bibinfo{person}{Wenbing Huang}, {and} \bibinfo{person}{Junzhou Huang}.} \bibinfo{year}{2020}\natexlab{}.
\newblock \showarticletitle{Self-Supervised Graph Transformer on Large-Scale Molecular Data}.
\newblock \bibinfo{journal}{\emph{Advances in Neural Information Processing Systems}}  \bibinfo{volume}{33} (\bibinfo{year}{2020}).
\newblock


\bibitem[\protect\citeauthoryear{Roziere, Gehring, Gloeckle, Sootla, Gat, Tan, Adi, Liu, Remez, Rapin, et~al\mbox{.}}{Roziere et~al\mbox{.}}{2023}]%
        {roziere2023code}
\bibfield{author}{\bibinfo{person}{Baptiste Roziere}, \bibinfo{person}{Jonas Gehring}, \bibinfo{person}{Fabian Gloeckle}, \bibinfo{person}{Sten Sootla}, \bibinfo{person}{Itai Gat}, \bibinfo{person}{Xiaoqing~Ellen Tan}, \bibinfo{person}{Yossi Adi}, \bibinfo{person}{Jingyu Liu}, \bibinfo{person}{Tal Remez}, \bibinfo{person}{J{\'e}r{\'e}my Rapin}, {et~al\mbox{.}}} \bibinfo{year}{2023}\natexlab{}.
\newblock \showarticletitle{Code llama: Open foundation models for code}.
\newblock \bibinfo{journal}{\emph{arXiv preprint arXiv:2308.12950}} (\bibinfo{year}{2023}).
\newblock


\bibitem[\protect\citeauthoryear{Shi, Huang, Feng, Zhong, Wang, and Sun}{Shi et~al\mbox{.}}{2021}]%
        {shi2020masked}
\bibfield{author}{\bibinfo{person}{Yunsheng Shi}, \bibinfo{person}{Zhengjie Huang}, \bibinfo{person}{Shikun Feng}, \bibinfo{person}{Hui Zhong}, \bibinfo{person}{Wenjin Wang}, {and} \bibinfo{person}{Yu Sun}.} \bibinfo{year}{2021}\natexlab{}.
\newblock \showarticletitle{Masked label prediction: Unified message passing model for semi-supervised classification}.
\newblock \bibinfo{journal}{\emph{IJCAI}} (\bibinfo{year}{2021}).
\newblock


\bibitem[\protect\citeauthoryear{Sohrabizadeh, Bai, Sun, and Cong}{Sohrabizadeh et~al\mbox{.}}{2022a}]%
        {sohrabizadeh2022automated}
\bibfield{author}{\bibinfo{person}{Atefeh Sohrabizadeh}, \bibinfo{person}{Yunsheng Bai}, \bibinfo{person}{Yizhou Sun}, {and} \bibinfo{person}{Jason Cong}.} \bibinfo{year}{2022}\natexlab{a}.
\newblock \showarticletitle{Automated accelerator optimization aided by graph neural networks}. In \bibinfo{booktitle}{\emph{Proceedings of the 59th ACM/IEEE Design Automation Conference}}. \bibinfo{pages}{55--60}.
\newblock


\bibitem[\protect\citeauthoryear{Sohrabizadeh, Bai, Sun, and Cong}{Sohrabizadeh et~al\mbox{.}}{2023}]%
        {sohrabizadeh2023robust}
\bibfield{author}{\bibinfo{person}{Atefeh Sohrabizadeh}, \bibinfo{person}{Yunsheng Bai}, \bibinfo{person}{Yizhou Sun}, {and} \bibinfo{person}{Jason Cong}.} \bibinfo{year}{2023}\natexlab{}.
\newblock \showarticletitle{Robust GNN-Based Representation Learning for HLS}. In \bibinfo{booktitle}{\emph{2023 IEEE/ACM International Conference on Computer Aided Design (ICCAD)}}. IEEE, \bibinfo{pages}{1--9}.
\newblock


\bibitem[\protect\citeauthoryear{Sohrabizadeh, Yu, Gao, and Cong}{Sohrabizadeh et~al\mbox{.}}{2022b}]%
        {autodse}
\bibfield{author}{\bibinfo{person}{Atefeh Sohrabizadeh}, \bibinfo{person}{Cody~Hao Yu}, \bibinfo{person}{Min Gao}, {and} \bibinfo{person}{Jason Cong}.} \bibinfo{year}{2022}\natexlab{b}.
\newblock \showarticletitle{AutoDSE: Enabling Software Programmers to Design Efficient FPGA Accelerators}.
\newblock \bibinfo{journal}{\emph{ACM Transactions on Design Automation of Electronic Systems (TODAES)}} \bibinfo{volume}{27}, \bibinfo{number}{4} (\bibinfo{year}{2022}), \bibinfo{pages}{1--27}.
\newblock


\bibitem[\protect\citeauthoryear{Sun, Hoffman, Verma, and Tang}{Sun et~al\mbox{.}}{2019}]%
        {sun2019infograph}
\bibfield{author}{\bibinfo{person}{Fan-Yun Sun}, \bibinfo{person}{Jordan Hoffman}, \bibinfo{person}{Vikas Verma}, {and} \bibinfo{person}{Jian Tang}.} \bibinfo{year}{2019}\natexlab{}.
\newblock \showarticletitle{InfoGraph: Unsupervised and Semi-supervised Graph-Level Representation Learning via Mutual Information Maximization}. In \bibinfo{booktitle}{\emph{International Conference on Learning Representations}}.
\newblock


\bibitem[\protect\citeauthoryear{Sun, Lin, and Zhu}{Sun et~al\mbox{.}}{2020}]%
        {sun2020multi}
\bibfield{author}{\bibinfo{person}{Ke Sun}, \bibinfo{person}{Zhouchen Lin}, {and} \bibinfo{person}{Zhanxing Zhu}.} \bibinfo{year}{2020}\natexlab{}.
\newblock \showarticletitle{Multi-Stage Self-Supervised Learning for Graph Convolutional Networks on Graphs with Few Labeled Nodes.}. In \bibinfo{booktitle}{\emph{AAAI}}. \bibinfo{pages}{5892--5899}.
\newblock


\bibitem[\protect\citeauthoryear{Svyatkovskiy, Deng, Fu, and Sundaresan}{Svyatkovskiy et~al\mbox{.}}{2020}]%
        {svyatkovskiy2020intellicode}
\bibfield{author}{\bibinfo{person}{Alexey Svyatkovskiy}, \bibinfo{person}{Shao~Kun Deng}, \bibinfo{person}{Shengyu Fu}, {and} \bibinfo{person}{Neel Sundaresan}.} \bibinfo{year}{2020}\natexlab{}.
\newblock \showarticletitle{Intellicode compose: Code generation using transformer}. In \bibinfo{booktitle}{\emph{Proceedings of the 28th ACM Joint Meeting on European Software Engineering Conference and Symposium on the Foundations of Software Engineering}}. \bibinfo{pages}{1433--1443}.
\newblock


\bibitem[\protect\citeauthoryear{Ustun, Deng, Pal, Li, and Zhang}{Ustun et~al\mbox{.}}{2020}]%
        {ustun2020accurate}
\bibfield{author}{\bibinfo{person}{Ecenur Ustun}, \bibinfo{person}{Chenhui Deng}, \bibinfo{person}{Debjit Pal}, \bibinfo{person}{Zhijing Li}, {and} \bibinfo{person}{Zhiru Zhang}.} \bibinfo{year}{2020}\natexlab{}.
\newblock \showarticletitle{Accurate operation delay prediction for FPGA HLS using graph neural networks}. In \bibinfo{booktitle}{\emph{Proceedings of the 39th International Conference on Computer-Aided Design}}. \bibinfo{pages}{1--9}.
\newblock


\bibitem[\protect\citeauthoryear{Vasudevan, Jiang, Bieber, Singh, Ho, Sutton, et~al\mbox{.}}{Vasudevan et~al\mbox{.}}{2021}]%
        {vasudevan2021learning}
\bibfield{author}{\bibinfo{person}{Shobha Vasudevan}, \bibinfo{person}{Wenjie~Joe Jiang}, \bibinfo{person}{David Bieber}, \bibinfo{person}{Rishabh Singh}, \bibinfo{person}{C~Richard Ho}, \bibinfo{person}{Charles Sutton}, {et~al\mbox{.}}} \bibinfo{year}{2021}\natexlab{}.
\newblock \showarticletitle{Learning semantic representations to verify hardware designs}.
\newblock \bibinfo{journal}{\emph{NeurIPS}}  \bibinfo{volume}{34} (\bibinfo{year}{2021}), \bibinfo{pages}{23491--23504}.
\newblock


\bibitem[\protect\citeauthoryear{Veli{\v{c}}kovi{\'{c}}, Fedus, Hamilton, Li{\`{o}}, Bengio, and Hjelm}{Veli{\v{c}}kovi{\'{c}} et~al\mbox{.}}{2019}]%
        {velickovic2018deep}
\bibfield{author}{\bibinfo{person}{Petar Veli{\v{c}}kovi{\'{c}}}, \bibinfo{person}{William Fedus}, \bibinfo{person}{William~L. Hamilton}, \bibinfo{person}{Pietro Li{\`{o}}}, \bibinfo{person}{Yoshua Bengio}, {and} \bibinfo{person}{R~Devon Hjelm}.} \bibinfo{year}{2019}\natexlab{}.
\newblock \showarticletitle{{Deep Graph Infomax}}. In \bibinfo{booktitle}{\emph{International Conference on Learning Representations}}.
\newblock
\urldef\tempurl%
\url{https://openreview.net/forum?id=rklz9iAcKQ}
\showURL{%
\tempurl}


\bibitem[\protect\citeauthoryear{Wang, Wang, Yang, Shen, Sun, Lee, and Han}{Wang et~al\mbox{.}}{2020}]%
        {wang2020gcn}
\bibfield{author}{\bibinfo{person}{Hanrui Wang}, \bibinfo{person}{Kuan Wang}, \bibinfo{person}{Jiacheng Yang}, \bibinfo{person}{Linxiao Shen}, \bibinfo{person}{Nan Sun}, \bibinfo{person}{Hae-Seung Lee}, {and} \bibinfo{person}{Song Han}.} \bibinfo{year}{2020}\natexlab{}.
\newblock \showarticletitle{GCN-RL circuit designer: Transferable transistor sizing with graph neural networks and reinforcement learning}. In \bibinfo{booktitle}{\emph{2020 57th ACM/IEEE Design Automation Conference (DAC)}}. IEEE, \bibinfo{pages}{1--6}.
\newblock


\bibitem[\protect\citeauthoryear{Wang, Wu, Yang, Hao, and Li}{Wang et~al\mbox{.}}{2022}]%
        {wang2022unsupervised}
\bibfield{author}{\bibinfo{person}{Haoyu~Peter Wang}, \bibinfo{person}{Nan Wu}, \bibinfo{person}{Hang Yang}, \bibinfo{person}{Cong Hao}, {and} \bibinfo{person}{Pan Li}.} \bibinfo{year}{2022}\natexlab{}.
\newblock \showarticletitle{Unsupervised Learning for Combinatorial Optimization with Principled Objective Relaxation}. In \bibinfo{booktitle}{\emph{NeurIPS}}.
\newblock


\bibitem[\protect\citeauthoryear{Wang, Wang, Joty, and Hoi}{Wang et~al\mbox{.}}{2021}]%
        {wang2021codet5}
\bibfield{author}{\bibinfo{person}{Yue Wang}, \bibinfo{person}{Weishi Wang}, \bibinfo{person}{Shafiq Joty}, {and} \bibinfo{person}{Steven~CH Hoi}.} \bibinfo{year}{2021}\natexlab{}.
\newblock \showarticletitle{Codet5: Identifier-aware unified pre-trained encoder-decoder models for code understanding and generation}.
\newblock \bibinfo{journal}{\emph{EMNLP}} (\bibinfo{year}{2021}).
\newblock


\bibitem[\protect\citeauthoryear{Wu, Xie, and Hao}{Wu et~al\mbox{.}}{2022a}]%
        {wu2022ironman}
\bibfield{author}{\bibinfo{person}{Nan Wu}, \bibinfo{person}{Yuan Xie}, {and} \bibinfo{person}{Cong Hao}.} \bibinfo{year}{2022}\natexlab{a}.
\newblock \showarticletitle{Ironman-pro: Multi-objective design space exploration in hls via reinforcement learning and graph neural network based modeling}.
\newblock \bibinfo{journal}{\emph{IEEE Transactions on Computer-Aided Design of Integrated Circuits and Systems}} (\bibinfo{year}{2022}).
\newblock


\bibitem[\protect\citeauthoryear{Wu, Yang, Xie, Li, and Hao}{Wu et~al\mbox{.}}{2022b}]%
        {wu2022high}
\bibfield{author}{\bibinfo{person}{Nan Wu}, \bibinfo{person}{Hang Yang}, \bibinfo{person}{Yuan Xie}, \bibinfo{person}{Pan Li}, {and} \bibinfo{person}{Cong Hao}.} \bibinfo{year}{2022}\natexlab{b}.
\newblock \showarticletitle{High-level synthesis performance prediction using gnns: Benchmarking, modeling, and advancing}. In \bibinfo{booktitle}{\emph{Proceedings of the 59th ACM/IEEE Design Automation Conference}}. \bibinfo{pages}{49--54}.
\newblock


\bibitem[\protect\citeauthoryear{Wu, He, Tang, Lv, and Liu}{Wu et~al\mbox{.}}{2021}]%
        {wu2021hanet}
\bibfield{author}{\bibinfo{person}{Peng Wu}, \bibinfo{person}{Xiangteng He}, \bibinfo{person}{Mingqian Tang}, \bibinfo{person}{Yiliang Lv}, {and} \bibinfo{person}{Jing Liu}.} \bibinfo{year}{2021}\natexlab{}.
\newblock \showarticletitle{Hanet: Hierarchical alignment networks for video-text retrieval}. In \bibinfo{booktitle}{\emph{Proceedings of the 29th ACM international conference on Multimedia}}. \bibinfo{pages}{3518--3527}.
\newblock


\bibitem[\protect\citeauthoryear{Xiao, Tian, Chen, Han, and Lewis}{Xiao et~al\mbox{.}}{2023}]%
        {xiao2023efficient}
\bibfield{author}{\bibinfo{person}{Guangxuan Xiao}, \bibinfo{person}{Yuandong Tian}, \bibinfo{person}{Beidi Chen}, \bibinfo{person}{Song Han}, {and} \bibinfo{person}{Mike Lewis}.} \bibinfo{year}{2023}\natexlab{}.
\newblock \showarticletitle{Efficient streaming language models with attention sinks}.
\newblock \bibinfo{journal}{\emph{arXiv preprint arXiv:2309.17453}} (\bibinfo{year}{2023}).
\newblock


\bibitem[\protect\citeauthoryear{Xie, Xu, Zhang, Wang, and Ji}{Xie et~al\mbox{.}}{2023}]%
        {xie2023self}
\bibfield{author}{\bibinfo{person}{Yaochen Xie}, \bibinfo{person}{Zhao Xu}, \bibinfo{person}{Jingtun Zhang}, \bibinfo{person}{Zhengyang Wang}, {and} \bibinfo{person}{Shuiwang Ji}.} \bibinfo{year}{2023}\natexlab{}.
\newblock \showarticletitle{Self-Supervised Learning of Graph Neural Networks: {A} Unified Review}.
\newblock \bibinfo{journal}{\emph{{IEEE} Trans. Pattern Anal. Mach. Intell.}} \bibinfo{volume}{45}, \bibinfo{number}{2} (\bibinfo{year}{2023}), \bibinfo{pages}{2412--2429}.
\newblock
\urldef\tempurl%
\url{https://doi.org/10.1109/TPAMI.2022.3170559}
\showDOI{\tempurl}


\bibitem[\protect\citeauthoryear{Xu, Li, Tian, Sonobe, Kawarabayashi, and Jegelka}{Xu et~al\mbox{.}}{2018}]%
        {xu2018representation}
\bibfield{author}{\bibinfo{person}{Keyulu Xu}, \bibinfo{person}{Chengtao Li}, \bibinfo{person}{Yonglong Tian}, \bibinfo{person}{Tomohiro Sonobe}, \bibinfo{person}{Ken-ichi Kawarabayashi}, {and} \bibinfo{person}{Stefanie Jegelka}.} \bibinfo{year}{2018}\natexlab{}.
\newblock \showarticletitle{Representation learning on graphs with jumping knowledge networks}.
\newblock \bibinfo{journal}{\emph{ICML}} (\bibinfo{year}{2018}).
\newblock


\bibitem[\protect\citeauthoryear{Xu, Salado, and Xie}{Xu et~al\mbox{.}}{2020}]%
        {xu2020reinforcement}
\bibfield{author}{\bibinfo{person}{Peng Xu}, \bibinfo{person}{Alejandro Salado}, {and} \bibinfo{person}{Guangrui Xie}.} \bibinfo{year}{2020}\natexlab{}.
\newblock \showarticletitle{A reinforcement learning approach to design verification strategies of engineered systems}. In \bibinfo{booktitle}{\emph{2020 IEEE International Conference on Systems, Man, and Cybernetics (SMC)}}. IEEE, \bibinfo{pages}{3543--3550}.
\newblock


\bibitem[\protect\citeauthoryear{Yang, He, and Cao}{Yang et~al\mbox{.}}{2022}]%
        {yang2022pre}
\bibfield{author}{\bibinfo{person}{Tai Yang}, \bibinfo{person}{Guoqing He}, {and} \bibinfo{person}{Peng Cao}.} \bibinfo{year}{2022}\natexlab{}.
\newblock \showarticletitle{Pre-routing path delay estimation based on transformer and residual framework}. In \bibinfo{booktitle}{\emph{2022 27th Asia and South Pacific Design Automation Conference (ASP-DAC)}}. IEEE, \bibinfo{pages}{184--189}.
\newblock


\bibitem[\protect\citeauthoryear{Yasunaga, Bosselut, Ren, Zhang, Manning, Liang, and Leskovec}{Yasunaga et~al\mbox{.}}{2022}]%
        {yasunaga2022deep}
\bibfield{author}{\bibinfo{person}{Michihiro Yasunaga}, \bibinfo{person}{Antoine Bosselut}, \bibinfo{person}{Hongyu Ren}, \bibinfo{person}{Xikun Zhang}, \bibinfo{person}{Christopher~D Manning}, \bibinfo{person}{Percy~S Liang}, {and} \bibinfo{person}{Jure Leskovec}.} \bibinfo{year}{2022}\natexlab{}.
\newblock \showarticletitle{Deep bidirectional language-knowledge graph pretraining}.
\newblock \bibinfo{journal}{\emph{Advances in Neural Information Processing Systems}}  \bibinfo{volume}{35} (\bibinfo{year}{2022}), \bibinfo{pages}{37309--37323}.
\newblock


\bibitem[\protect\citeauthoryear{You, Chen, Sui, Chen, Wang, and Shen}{You et~al\mbox{.}}{2020}]%
        {You2020GraphCL}
\bibfield{author}{\bibinfo{person}{Yuning You}, \bibinfo{person}{Tianlong Chen}, \bibinfo{person}{Yongduo Sui}, \bibinfo{person}{Ting Chen}, \bibinfo{person}{Zhangyang Wang}, {and} \bibinfo{person}{Yang Shen}.} \bibinfo{year}{2020}\natexlab{}.
\newblock \showarticletitle{Graph Contrastive Learning with Augmentations}. In \bibinfo{booktitle}{\emph{Advances in Neural Information Processing Systems}}, \bibfield{editor}{\bibinfo{person}{H.~Larochelle}, \bibinfo{person}{M.~Ranzato}, \bibinfo{person}{R.~Hadsell}, \bibinfo{person}{M.~F. Balcan}, {and} \bibinfo{person}{H.~Lin}} (Eds.), Vol.~\bibinfo{volume}{33}. \bibinfo{publisher}{Curran Associates, Inc.}, \bibinfo{pages}{5812--5823}.
\newblock
\urldef\tempurl%
\url{https://proceedings.neurips.cc/paper/2020/file/3fe230348e9a12c13120749e3f9fa4cd-Paper.pdf}
\showURL{%
\tempurl}


\bibitem[\protect\citeauthoryear{Yuki and Pouchet}{Yuki and Pouchet}{[n.d.]}]%
        {polybench}
\bibfield{author}{\bibinfo{person}{Tomofumi Yuki} {and} \bibinfo{person}{Louis-Noël Pouchet}.} \bibinfo{year}{[n.d.]}\natexlab{}.
\newblock \bibinfo{title}{PolyBench/C}.
\newblock
\newblock
\urldef\tempurl%
\url{https://web.cse.ohio-state.edu/~pouchet.2/software/polybench/}
\showURL{%
\tempurl}


\bibitem[\protect\citeauthoryear{Zhang, Bosselut, Yasunaga, Ren, Liang, Manning, and Leskovec}{Zhang et~al\mbox{.}}{2022}]%
        {zhang2022greaselm}
\bibfield{author}{\bibinfo{person}{Xikun Zhang}, \bibinfo{person}{Antoine Bosselut}, \bibinfo{person}{Michihiro Yasunaga}, \bibinfo{person}{Hongyu Ren}, \bibinfo{person}{Percy Liang}, \bibinfo{person}{Christopher~D Manning}, {and} \bibinfo{person}{Jure Leskovec}.} \bibinfo{year}{2022}\natexlab{}.
\newblock \showarticletitle{Greaselm: Graph reasoning enhanced language models}. In \bibinfo{booktitle}{\emph{International conference on learning representations}}.
\newblock


\bibitem[\protect\citeauthoryear{Zhang, Han, Liu, Jiang, Sun, and Liu}{Zhang et~al\mbox{.}}{2019}]%
        {zhang2019ernie}
\bibfield{author}{\bibinfo{person}{Zhengyan Zhang}, \bibinfo{person}{Xu Han}, \bibinfo{person}{Zhiyuan Liu}, \bibinfo{person}{Xin Jiang}, \bibinfo{person}{Maosong Sun}, {and} \bibinfo{person}{Qun Liu}.} \bibinfo{year}{2019}\natexlab{}.
\newblock \showarticletitle{ERNIE: Enhanced language representation with informative entities}.
\newblock \bibinfo{journal}{\emph{ACL}} (\bibinfo{year}{2019}).
\newblock


\bibitem[\protect\citeauthoryear{Zheng, Xia, Zou, Dong, Wang, Xue, Wang, Shen, Wang, Li, et~al\mbox{.}}{Zheng et~al\mbox{.}}{2023}]%
        {zheng2023codegeex}
\bibfield{author}{\bibinfo{person}{Qinkai Zheng}, \bibinfo{person}{Xiao Xia}, \bibinfo{person}{Xu Zou}, \bibinfo{person}{Yuxiao Dong}, \bibinfo{person}{Shan Wang}, \bibinfo{person}{Yufei Xue}, \bibinfo{person}{Zihan Wang}, \bibinfo{person}{Lei Shen}, \bibinfo{person}{Andi Wang}, \bibinfo{person}{Yang Li}, {et~al\mbox{.}}} \bibinfo{year}{2023}\natexlab{}.
\newblock \showarticletitle{Codegeex: A pre-trained model for code generation with multilingual evaluations on humaneval-x}.
\newblock \bibinfo{journal}{\emph{arXiv preprint arXiv:2303.17568}} (\bibinfo{year}{2023}).
\newblock


\end{thebibliography}

\clearpage
\appendix
\subsection{Additional Background in HLS Design}\label{app:hls}

\small
\vspace{0.12in}
\begin{lstlisting}[language=C,caption={\small{Code snippet of the \textsc{mvt} kernel (Matrix Vector Product and Transpose) with its 8 pragmas starting with ``\texttt{\#pragma}''. 
% Setting pragmas to different parameters leads to different designs with unique microarchitectures for which we aim to predict the quality in terms of latency and resource utilization on FPGA.
}},label={code:mvt},floatplacement=H]
void kernel_mvt(double x1[120], double x2[120], double y_1[120], double y_2[120], double A[120][120]) {
  int i, j;
#pragma ACCEL PIPELINE auto{__PIPE__L0}
#pragma ACCEL TILE FACTOR=auto{__TILE__L0}
#pragma ACCEL PARALLEL FACTOR=auto{__PARA__L0}
  for (i = 0; i < 120; i++) {
#pragma ACCEL PARALLEL reduction = x1 FACTOR=auto{__PARA__L2}
    for (j = 0; j < 120; j++) {
      x1[i] += A[i][j] * y_1[j];
    }}
#pragma ACCEL PIPELINE auto{__PIPE__L1}
#pragma ACCEL TILE FACTOR=auto{__TILE__L1}
#pragma ACCEL PARALLEL FACTOR=auto{__PARA__L1}
  for (i = 0; i < 120; i++) { 
#pragma ACCEL PARALLEL reduction = x2 FACTOR=auto{__PARA__L3}
    for (j = 0; j < 120; j++) {
      x2[i] += A[j][i] * y_2[j];
    }}}
\end{lstlisting}
\normalsize

\subsection{Extra Embedding Visualization}

\begin{figure}[htbp]
    \centering
    \begin{subfigure}[b]{0.23\textwidth}
        \centering
        \includegraphics[width=\textwidth,height=0.12\textheight]{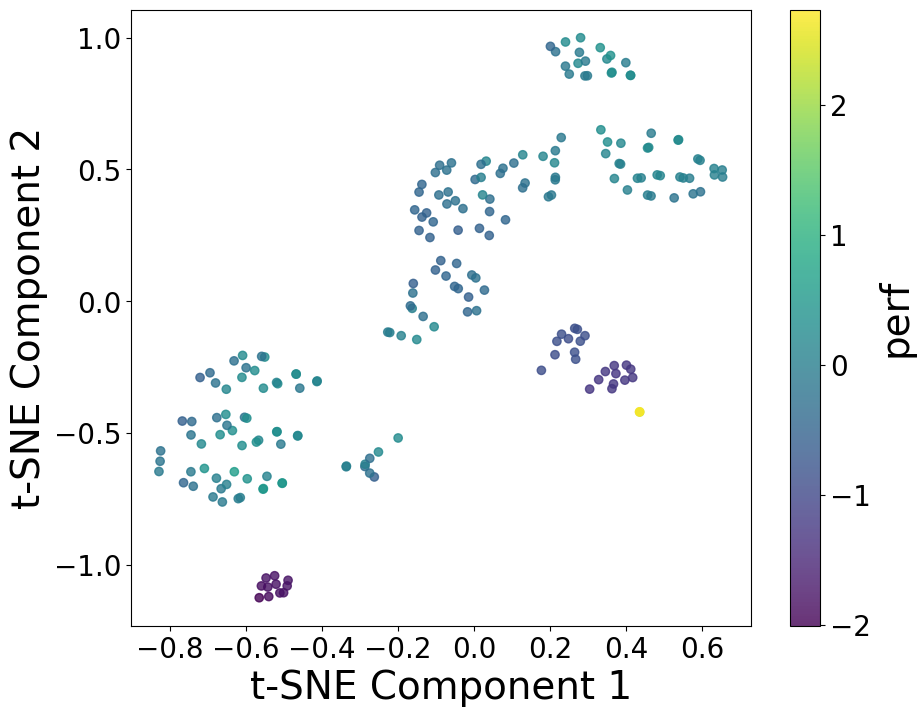}
        \caption{HARP embeddings}
        \label{fig:harp_vis}
    \end{subfigure}
    \hfill
    \begin{subfigure}[b]{0.23\textwidth}
        \centering
        \includegraphics[width=\textwidth,height=0.12\textheight]{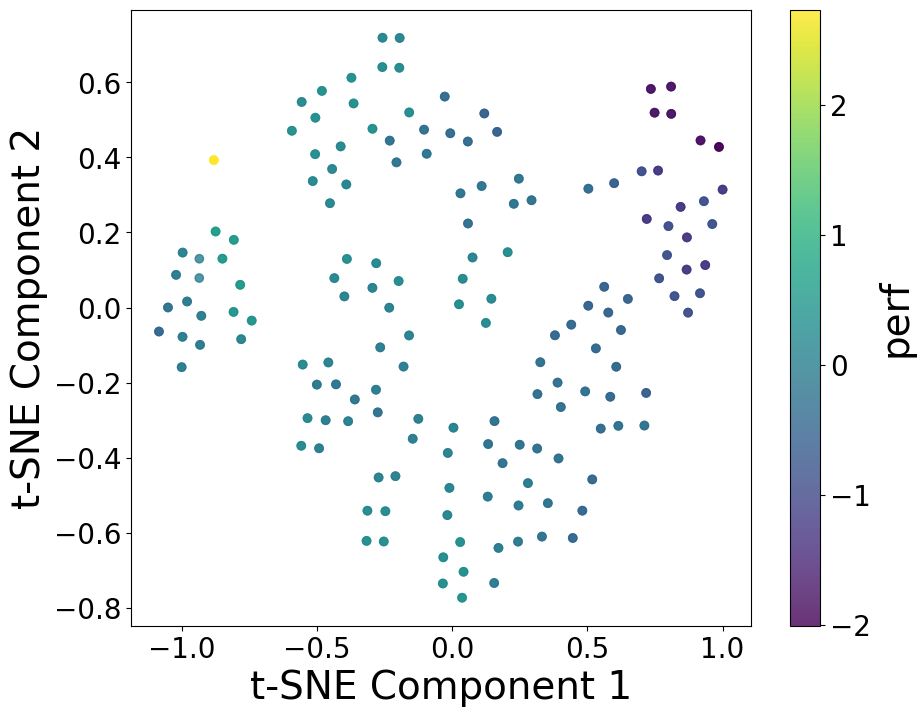}
        \caption{CodeT5 embeddings}
        \label{fig:codet5-vis}
    \end{subfigure}
    \hfill
    \begin{subfigure}[b]{0.23\textwidth}
        \centering
        \includegraphics[width=\textwidth,height=0.12\textheight]{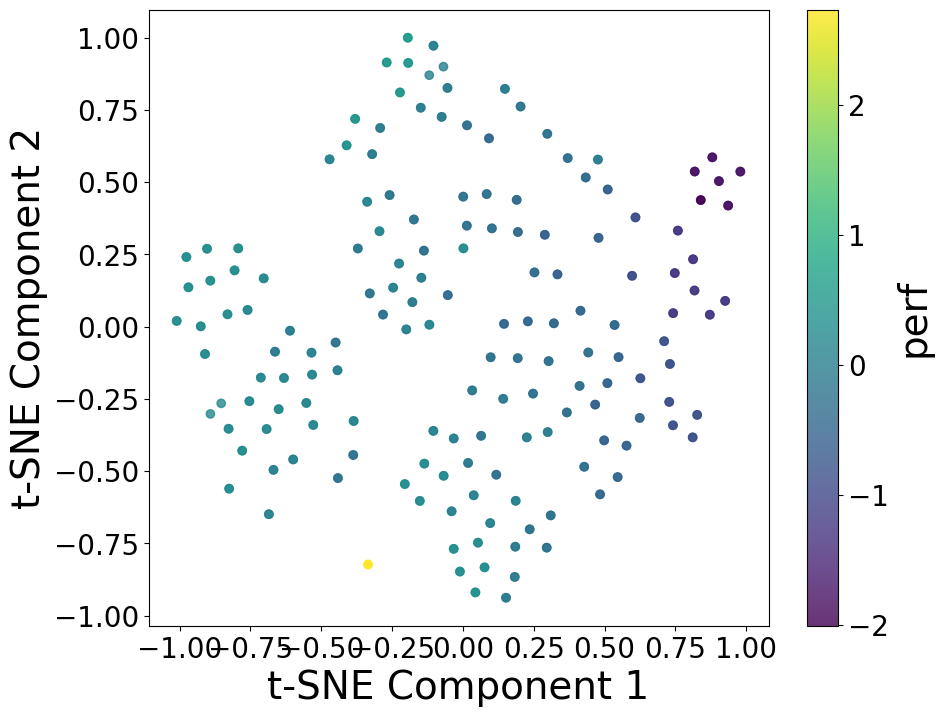}
        \caption{ProgSG embeddings}
        \label{fig:progsg-vis}
    \end{subfigure}
    \caption{Embedding visualizations with different methods for ``\textit{Symm-Opt-Medium kernel}'' using t-SNE. The color indicates the value of ``perf'' target.}
    \label{fig:emb_vis}
\end{figure}

In Figure \ref{fig:emb_vis} we visualize the embeddings of different models for ``\textit{symm-opt-medium}'' kernel, where \model achieves more than eight times speed up compared to \cdfg. Similar to the embeddings of ``Correlation'' kernel, the embeddings of \cdfg are more crowded, suggesting weaker generalization ability. Moreover, comparing the embeddings of \codetfive and \model, we can see that the yellow point (which represents the design point with the best performance) in \model's embeddings are further away from other points than in \codetfive's embeddings. It suggests \model can better distinguish the good design points. 

\subsection{Case Studies of Best Design Points in DSE experiments}

In this section, we show the design points returned by AutoDSE (25 hours), \cdfg (1 hour), and \model (1 hour) in the DSE experiments for some kernels (``Correlation'', ``Symm-opt-medium'', ``Gemver-medium'') where \model outperforms \cdfg significantly. We show the source code of these kernels in Code \ref{code:hls-cor}, Code \ref{code:hls-symm}, and Code \ref{code:hls-gemver}. And we show the design points in Table \ref{table:corr}, Table \ref{table:symm}, and Table \ref{table:gemver}.

For \textit{correlation} kernel, the values of ``\_\_PARA\_\_L5'' in AutoDSE and \model's design points are much larger than the value in \cdfg's design point. So the design point returned by \cdfg leads to sub-optimal data loading procedures, which takes up 96,000 cycles of the total latency. While the design point returned by \model does not have this issue. For \textit{symm-opt-medium} kernel, the design point returned by \cdfg has smaller parallelization factor than the design points returned by AutoDSE and \model, leading to worse efficiency. For \textit{gemver-medium} kernel, the parameter ``\_\_PARA\_\_L4'' is 64 in the design point returned by \cdfg, which can not divide 400, which is the total number of the for-loop. As the result, the loop takes up 120,000 cycles of the total latency, leading to worse performance. On the other hand, The ``\_\_PARA\_\_L4'' is 25 in the design point returned by \model, which can divide 400, thus avoiding the problem.

\vspace{3mm}
{
\small
\begin{lstlisting}[language=C,caption={\small{Code snippet of the \textsc{Correlation} kernel with its pragmas starting with ``\texttt{\#pragma}''. 
% Setting pragmas to different parameters leads to different designs with unique microarchitectures for which we aim to predict the quality in terms of latency and resource utilization on FPGA.
}},label={code:hls-cor},floatplacement=H]
void kernel_correlation(double float_n,double data[100][80],double corr[80][80],double mean[80],double stddev[80])
{
  int i;
  int j;
  int k;
  double eps = 0.1;
  
#pragma ACCEL PIPELINE auto{__PIPE__L0}
#pragma ACCEL TILE FACTOR=auto{__TILE__L0}
#pragma ACCEL PARALLEL FACTOR=auto{__PARA__L0}
  for (j = 0; j < 80; j++) {
    mean[j] = 0.0;
#pragma ACCEL PARALLEL FACTOR=auto{__PARA__L4}
    for (i = 0; i < 100; i++) {
      mean[j] += data[i][j];
    }
    mean[j] /= float_n;
  }
#pragma ACCEL PIPELINE auto{__PIPE__L1}
#pragma ACCEL TILE FACTOR=auto{__TILE__L1}
#pragma ACCEL PARALLEL FACTOR=auto{__PARA__L1}
  for (j = 0; j < 80; j++) {
    stddev[j] = 0.0;
#pragma ACCEL PARALLEL FACTOR=auto{__PARA__L5}
    for (i = 0; i < 100; i++) {
      stddev[j] += pow(data[i][j] - mean[j],(double )2);
    }
    stddev[j] /= float_n;
    stddev[j] = sqrt(stddev[j]);
    stddev[j] = (stddev[j] <= eps?1.0 : stddev[j]);
  }
#pragma ACCEL PIPELINE auto{__PIPE__L2}
#pragma ACCEL TILE FACTOR=auto{__TILE__L2}
#pragma ACCEL PARALLEL FACTOR=auto{__PARA__L2}
  for (i = 0; i < 100; i++) {
#pragma ACCEL PARALLEL FACTOR=auto{__PARA__L6}
    for (j = 0; j < 80; j++) {
      data[i][j] -= mean[j];
      data[i][j] /= sqrt(float_n) * stddev[j];
    }
  }
#pragma ACCEL PIPELINE auto{__PIPE__L3}
#pragma ACCEL TILE FACTOR=auto{__TILE__L3}
#pragma ACCEL PARALLEL FACTOR=auto{__PARA__L3}
  for (i = 0; i < 80 - 1; i++) {
    corr[i][i] = 1.0;  
#pragma ACCEL PIPELINE auto{__PIPE__L7}
    for (j = i + 1; j < 80; j++) {
      corr[i][j] = 0.0;
#pragma ACCEL PARALLEL FACTOR=auto{__PARA__L7_0}
      for (k = 0; k < 100; k++) {
        corr[i][j] += data[k][i] * data[k][j];
      }
      corr[j][i] = corr[i][j];
    }
  }
  corr[80 - 1][80 - 1] = 1.0;
}
\end{lstlisting}
\normalsize
}

\begin{table}[h!]
\centering
\begin{tabular}{|>{\ttfamily}l|>{\ttfamily}l|>{\ttfamily}l|>{\ttfamily}l|}
\toprule
 & \textbf{AutoDSE} & \textbf{HARP} & \textbf{ProgSG} \\
\midrule
\_\_PARA\_\_L0  & 1       & 1       & 1       \\
\_\_PARA\_\_L1  & 1       & 1       & 1       \\
\_\_PARA\_\_L2  & 1       & 1       & 1       \\
\_\_PARA\_\_L3  & 1       & 1       & 1       \\
\_\_PARA\_\_L4  & 1       & 5       & 4       \\
\_\_PARA\_\_L5  & 32      & 5       & 25      \\
\_\_PARA\_\_L6  & 1       & 10      & 4       \\
\_\_PARA\_\_L7\_0 & 1     & 1       & 1       \\
\_\_PIPE\_\_L0  & fg & off     & off     \\
\_\_PIPE\_\_L1  & off     & off     & off     \\
\_\_PIPE\_\_L2  & off     & off     & off     \\
\_\_PIPE\_\_L3  & off     & off     & off     \\
\_\_PIPE\_\_L7  & fg & fg & fg \\
\_\_TILE\_\_L0  & 1       & 1       & 1       \\
\_\_TILE\_\_L1  & 1       & 1       & 1       \\
\_\_TILE\_\_L2  & 1       & 1       & 1       \\
\_\_TILE\_\_L3  & 1       & 1       & 1       \\\midrule
perf & 60,237 & 165,135 & 61,287\\
\bottomrule
\end{tabular}
\caption{Best design points returned by AutoDSE, HARP, and ProgSG on ``\textit{Correlation}'' kernel.}
\label{table:corr}
\end{table}

\vspace{3mm}
{
\small
\begin{lstlisting}[language=C,caption={\small{Code snippet of the \textsc{Symm-opt-medium} kernel with its pragmas starting with ``\texttt{\#pragma}''. 
% Setting pragmas to different parameters leads to different designs with unique microarchitectures for which we aim to predict the quality in terms of latency and resource utilization on FPGA.
}},label={code:hls-symm},floatplacement=H]
void kernel_symm(double alpha,double beta,double C[200][240],double A[200][200],double B[200][240])
{
  int i,j,k;
#pragma ACCEL PIPELINE auto{__PIPE__L0}
#pragma ACCEL TILE FACTOR=auto{__TILE__L0}
#pragma ACCEL PARALLEL FACTOR=auto{__PARA__L0}
  for (i = 0; i < 200; i++) {
#pragma ACCEL PIPELINE auto{__PIPE__L1}
#pragma ACCEL TILE FACTOR=auto{__TILE__L1}
#pragma ACCEL PARALLEL FACTOR=auto{__PARA__L1}
    for (j = 0; j < 240; j++) {
      double tmp = B[i][j];
#pragma ACCEL PARALLEL reduction=C FACTOR=auto{__PARA__L2}
      for (k = 0; k < 200; k++) {
        if (k < i) {
          C[k][j] += alpha * tmp * A[i][k];
        }
      }
      double temp2 = (double )0;
#pragma ACCEL PARALLEL reduction=temp2 FACTOR=auto{__PARA__L3}
      for (k = 0; k < 200; k++) {
        if (k < i) {
          temp2 += B[k][j] * A[i][k];
        }
      }
      C[i][j] = beta * C[i][j] + alpha * B[i][j] * A[i][i] + alpha * temp2;
    }
  }
}
\end{lstlisting}
\normalsize
}

\begin{table}[h!]
\centering
\begin{tabular}{|>{\ttfamily}l|>{\ttfamily}l|>{\ttfamily}l|>{\ttfamily}l|}
\toprule
 & \textbf{AutoDSE} & \textbf{HARP} & \textbf{ProgSG} \\
\midrule
\_\_PARA\_\_L0  & 1       & 1       & 1       \\
\_\_PARA\_\_L1  & 1       & 1       & 1       \\
\_\_PARA\_\_L2  & 25      & 25      & 25      \\
\_\_PARA\_\_L3  & 200     & 32      & 200     \\
\_\_PIPE\_\_L0  & cg      & off     & cg      \\
\_\_PIPE\_\_L1  & off     &   cg      & off     \\
\_\_TILE\_\_L0  & 1       & 1       & 1       \\
\_\_TILE\_\_L1  & 1       & 8       & 1       \\
\midrule
perf & 4,345,927 &35,536,546 & 4,345,927\\
\bottomrule
\end{tabular}
\caption{Best design points returned by AutoDSE, HARP, and ProgSG on ``\textit{Symm-OPT-Medium}'' kernel.}
\label{table:symm}
\end{table}

\vspace{3mm}
{
\small
\begin{lstlisting}[language=C,caption={\small{Code snippet of the \textsc{Gemver-medium} kernel with its pragmas starting with ``\texttt{\#pragma}''. 
% Setting pragmas to different parameters leads to different designs with unique microarchitectures for which we aim to predict the quality in terms of latency and resource utilization on FPGA.
}},label={code:hls-gemver},floatplacement=H]
void kernel_gemver(int n,double alpha,double beta,double A[400][400],double u1[400],double v1[400],double u2[400],double v2[400],double w[400],double x[400],double y[400],double z[400])
{
  int i,j;
#pragma ACCEL PIPELINE auto{__PIPE__L0}
#pragma ACCEL TILE FACTOR=auto{__TILE__L0}
#pragma ACCEL PARALLEL FACTOR=auto{__PARA__L0}
  for (i = 0; i < 400; i++) {
#pragma ACCEL PARALLEL reduction=A FACTOR=auto{__PARA__L4}
    for (j = 0; j < 400; j++) {
      A[i][j] += + u1[i] * v1[j] + u2[i] * v2[j];
    }
  }
#pragma ACCEL PIPELINE auto{__PIPE__L1}
#pragma ACCEL TILE FACTOR=auto{__TILE__L1}
#pragma ACCEL PARALLEL FACTOR=auto{__PARA__L1}
  for (i = 0; i < 400; i++) {
#pragma ACCEL PARALLEL reduction=x FACTOR=auto{__PARA__L5}
    for (j = 0; j < 400; j++) {
      x[i] += beta * A[j][i] * y[j];
    }
  }
#pragma ACCEL PARALLEL FACTOR=auto{__PARA__L2}
  for (i = 0; i < 400; i++) {
    x[i] = x[i] + z[i];
  }
#pragma ACCEL PIPELINE auto{__PIPE__L3}
#pragma ACCEL TILE FACTOR=auto{__TILE__L3}
#pragma ACCEL PARALLEL FACTOR=auto{__PARA__L3}
  for (i = 0; i < 400; i++) {
#pragma ACCEL PARALLEL reduction=w FACTOR=auto{__PARA__L6}
    for (j = 0; j < 400; j++) {
      w[i] += alpha * A[i][j] * x[j];
    }}}
\end{lstlisting}
\normalsize
}

\begin{table}[h!]
\centering
\begin{tabular}{|>{\ttfamily}l|>{\ttfamily}l|>{\ttfamily}l|>{\ttfamily}l|}
\toprule
& \textbf{AutoDSE} & \textbf{HARP} & \textbf{ProgSG} \\
\midrule
\_\_PARA\_\_L0  & 1       & 1       & 1       \\
\_\_PARA\_\_L1  & 1       & 1       & 1       \\
\_\_PARA\_\_L2  & 1       & 1       & 1       \\
\_\_PARA\_\_L3  & 1       & 8       & 8       \\
\_\_PARA\_\_L4  & 2       & 64      & 25      \\
\_\_PARA\_\_L5  & 1       & 10      & 10      \\
\_\_PARA\_\_L6  & 25      & 20      & 25      \\
\_\_PIPE\_\_L0  & off     & off     & off     \\
\_\_PIPE\_\_L1  & fg & off     & cg      \\
\_\_PIPE\_\_L3  & off     &  cg      & off     \\
\_\_TILE\_\_L0  & 1       & 1       & 1       \\
\_\_TILE\_\_L1  & 1       & 1       & 1       \\
\_\_TILE\_\_L3  & 8       & 1       & 1       \\
\midrule
perf & 210,335 & 265,686 & 167,270\\
\bottomrule
\end{tabular}
\caption{Best design points returned by AutoDSE, HARP, and ProgSG on ``\textit{Gemver-medium}'' kernel.}
\label{table:gemver}
\end{table}

\end{document}